\documentclass[11pt, a4paper, logo, copyright, nonumbering]{deepseek}
\usepackage[numbers, sort&compress]{natbib}

\usepackage{times}              % Times font
\usepackage{xspace}             % Smart spacing after commands
\usepackage{pifont}             % Special symbols and fonts
\usepackage{multirow}           % Multi-row cells in tables
\usepackage{tcolorbox}          % Colored boxes and frames
\usepackage{xltabular}          % Extended tabular with longtable features
\usepackage{longtable}          % Tables spanning multiple pages
\usepackage{hyperref}           % Clickable links and references
\usepackage{amsfonts}           % AMS fonts
\usepackage{amsmath}            % Enhanced math environments
\usepackage{amssymb}            % AMS symbols
\usepackage{lineno}             % Line numbering
\usepackage{adjustbox}          % Adjustable boxes for scaling content
\usepackage[bottom]{footmisc}   % Footnote positioning
\usepackage{CJKutf8}            % Chinese, Japanese, Korean support
\usepackage{subfigure}          % Subfigures (consider subfig or subcaption instead)
\usepackage{setspace}           % Line spacing control
\usepackage{dsfont}             % Double-stroke fonts (for mathematical sets)
\usepackage{array}              % Enhanced array and tabular
\usepackage{tabularx}           % Auto-sizing table columns
\usepackage{xcolor}             % Extended color support
\usepackage{booktabs}           % Professional table formatting
\usepackage{wrapfig}            % Text wrapping around figures
\usepackage{lipsum}             % Lorem ipsum text generation
\usepackage{multicol}           % Multiple columns
\usepackage{url}                % URL formatting
\usepackage{graphicx}           % Graphics inclusion
\usepackage{colortbl}           % Colored table cells
\usepackage{enumitem}           % Enhanced list formatting

\tcbuselibrary{breakable,skins}

\usepackage{fancyhdr}
\newtcolorbox{myboxi}[1][]{
  breakable,
  title=#1,
  colback=blue!5,
  colbacktitle=blue!5,
  coltitle=black,
  fonttitle=\bfseries,
  bottomrule=0pt,
  toprule=0pt,
  leftrule=2pt,
  rightrule=2pt,
  titlerule=0pt,
  arc=0pt,
  outer arc=0pt,
  colframe=blue,
}

\definecolor{HeaderBG}  {HTML}{B5CCE9}
\definecolor{CategoryBG}{HTML}{DDEBF7}
\definecolor{GREEN}{HTML}{E2F0D9}
\definecolor{AltGray}   {HTML}{F5F5F5}

\renewcommand{\arraystretch}{1.2}

\makeatletter
\def\@BTrule[#1]{%
  \ifx\longtable\undefined
    \let\@BTswitch\@BTnormal
  \else\ifx\hline\LT@hline
    \nobreak
    \let\@BTswitch\@BLTrule
  \else
     \let\@BTswitch\@BTnormal
  \fi\fi
  \global\@thisrulewidth=#1\relax
  \ifnum\@thisruleclass=\tw@\vskip\@aboverulesep\else
  \ifnum\@lastruleclass=\z@\vskip\@aboverulesep\else
  \ifnum\@lastruleclass=\@ne\vskip\doublerulesep\fi\fi\fi
  \@BTswitch}
\makeatother

\addto\extrasenglish{
}

 {\begin{list}{}%
         {\setlength{\leftmargin}{#1}}%
         \item[]%
 }
 {\end{list}}
 
\reportnumber{001}
\renewcommand{\today}{}

\title{\centering Real Deep Research for AI, Robotics and Beyond}
\renewcommand\Affilfont{\bfseries\fontsize{10}{12}\selectfont\centering}

\author[1]{Xueyan Zou$^*$}
\author[1]{Jianglong Ye$^*$} 
\author[2]{Hao Zhang}
\author[3]{Xiaoyu Xiang}
\author[5]{Mingyu Ding}
\author[1]{Zhaojing Yang} 
\author[4]{\\ Yong Jae Lee}
\author[1]{Zhuowen Tu}
\author[2]{Sifei Liu}
\author[1]{Xiaolong Wang}

\affil[1]{UC San Diego}
\affil[2]{NVIDIA}
\affil[3]{META}
\affil[4]{UW-Madison}
\affil[5]{UNC}

\affil[ ]{\hspace{1pt} {\color{blue}{\texttt{\url{https://realdeepresearch.github.io/}}}}}

\renewcommand{\phi}{\varphi}

\renewcommand{\epsilon}{\varepsilon}
\renewcommand{\imath}{\mathrm{i}}

\newlength{\restsubwidth}
\newlength{\restsubheight}
\newlength{\restsubmoreheight}
\newcommand{\rest}[2]{%
        \settowidth{\restsubwidth}{\ensuremath{#2}}
        \settoheight{\restsubheight}{\ensuremath{{}_{#2}}}
        \ensuremath{{#1\hskip 0.5pt}_{\vrule\kern2pt\parbox[b][%
        4pt][b]{\the\restsubwidth}{%
                        \ensuremath{{}_{#2}}}}}
        }

\definecolor{LightGreen}{rgb}{0.9, 1, 0.9}

\begin{document}
\begin{abstract}
\vspace{-5pt}
With the rapid growth of research in modern AI and robotics—now producing over 10,000 papers annually—it has become increasingly difficult for researchers to stay up to date. Fast-evolving trends, \textit{the rise of interdisciplinary work}, and \textit{the need to explore domains beyond one’s expertise} all contribute to this challenge. To address these issues, we propose a generalizable pipeline capable of systematically analyzing any research area: identifying emerging trends, uncovering cross-domain opportunities, and offering concrete starting points for new inquiry. In this work, we present \textbf{Real Deep Research} (RDR)—a comprehensive framework applied to the domains of AI and robotics, with a particular focus on foundation models and robotics advancements. We also briefly extend our analysis to other areas of science. The main paper details the construction of the RDR pipeline, while the appendix provides extensive results across each analyzed topic. We hope this work could shed lights on researchers who works in the filed of AI and beyond.
\vspace{5pt}
\end{abstract}
\maketitle

\begin{figure}[h!]
    \centering
    \includegraphics[width=1.\textwidth]{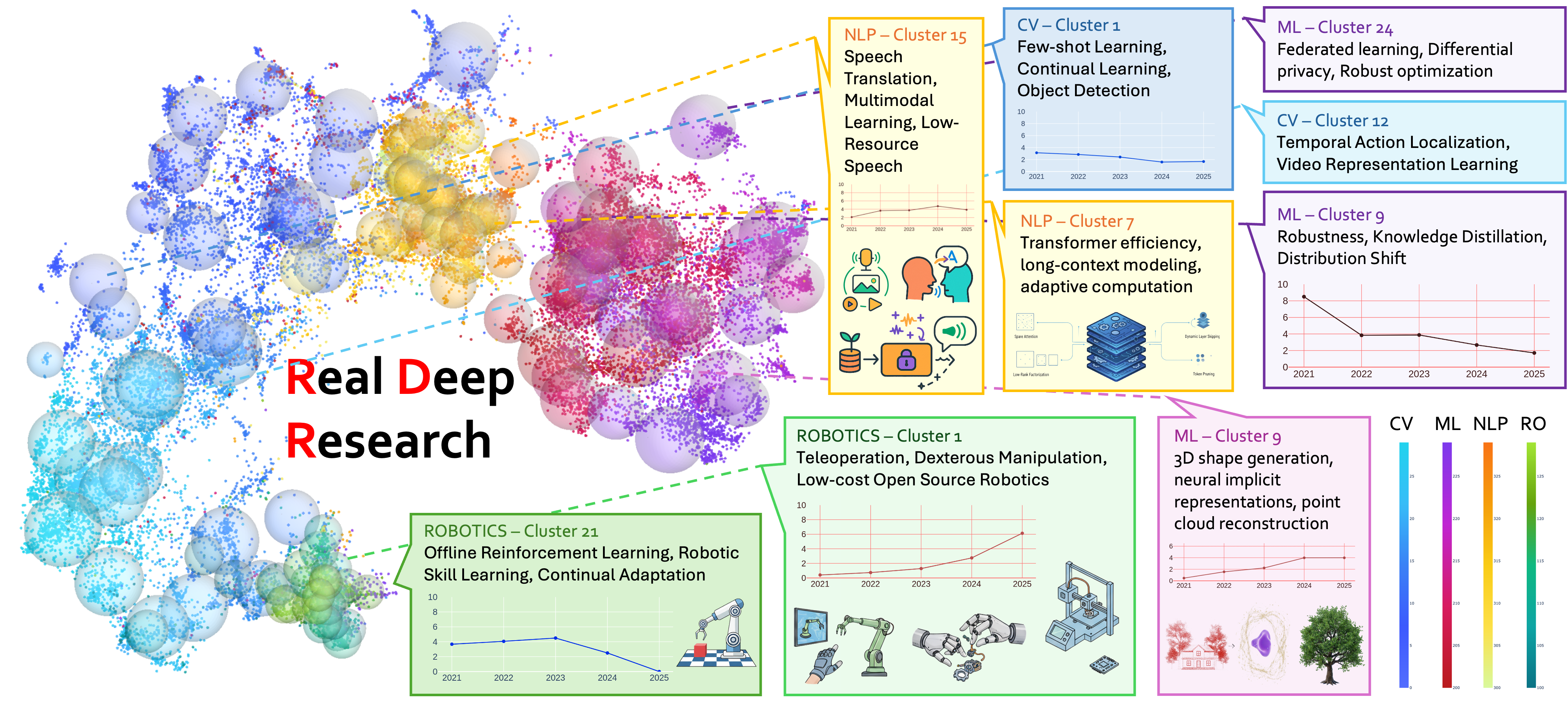}
    \caption{Real Deep Research enables: (1) generating surveys for specific research focuses or perspectives; (2) analyzing topic trends over time; (3) mapping interdisciplinary research landscapes; and (4) retrieving high-impact papers relevant to a given topic. (Each dot represents a paper, and each sphere denotes a topic cluster. The cluster keywords and trend information are automatically generated by \textbf{RDR}) \firstpagefoot{\textbf{*} Indicate core contribution.}}
    \label{fig:teaser}
\end{figure}

\begin{abstract}
\vspace{-5pt}
With the rapid growth of research in modern AI and robotics—now producing over 10,000 papers annually—it has become increasingly difficult for researchers to stay up to date. Fast-evolving trends, \textit{the rise of interdisciplinary work}, and \textit{the need to explore domains beyond one’s expertise} all contribute to this challenge. To address these issues, we propose a generalizable pipeline capable of systematically analyzing any research area: identifying emerging trends, uncovering cross-domain opportunities, and offering concrete starting points for new inquiry. In this work, we present \textbf{Real Deep Research} (RDR)—a comprehensive framework applied to the domains of AI and robotics, with a particular focus on foundation models and robotics advancements. We also briefly extend our analysis to other areas of science. The main paper details the construction of the RDR pipeline, while the appendix provides extensive results across each analyzed topic. We hope this work could shed lights on researchers who works in the filed of AI and beyond.
\vspace{5pt}
\end{abstract}
\vspace{-5pt}
\section{Introduction}

The fields of AI and robotics have experienced exponential growth in recent years, while researchers continue to face the constraint of limited time and attention. This work is motivated by the authors’ need to efficiently survey research areas, stay up to date with rapidly evolving trends, identify promising interdisciplinary opportunities, and familiarize themselves with the latest developments on a given topic.

In response to this need, we develop a systematic analysis tool designed to help users quickly navigate and adapt to any research area or topic. We begin by applying our approach to the fields of AI and robotics, conducting an in-depth analysis with a focus on foundation models and robotics research. To broaden our exploration and uncover emerging areas of interest, we also extend our analysis to natural sciences and formal sciences, offering a glimpse into recent developments beyond our core domains.

Although our intentions are well-founded, it is important to acknowledge existing efforts in this space. On the one hand, there are high-quality survey papers written by domain experts~\cite{bommasani2021opportunities,Huang2022TowardsRI}; on the other hand, a few recent works have explored automated research pipelines~\cite{ajith2024litsearch,li2024scilitllm}. Expert-written surveys offer depth and accuracy, but require significant manual effort and cannot easily adapt to the fast-paced evolution of research. Meanwhile, current automated approaches often lack domain-specific knowledge and expert insight, limiting their usefulness and relevance to researchers. Our work aims to bridge this gap by combining systematic automation with meaningful, expert-informed analysis.

Therefore, in addition to building an effective pipeline for Real Deep Research, our goal is to make the tool robust and insightful enough to support top-tier researchers in tracking emerging trends and engaging with unfamiliar research areas. A key focus of our work is interdisciplinary exploration—helping researchers identify underexplored intersections between fields that present promising opportunities for cross-domain collaboration.

As shown in Fig.~\ref{fig:teaser}, the visualization displays individual papers, clustered research topics, and their corresponding trends. At a glance, it becomes clear that areas such as teleoperation, dexterous manipulation, and open-source robotics are emerging as promising directions, whereas traditional reinforcement learning appears to be declining in momentum. As researchers in the robotics field, we find that these trend insights align well with our domain knowledge and provide valuable guidance for identifying impactful research opportunities. We summarize the key contributions of this paper as follows:

\begin{enumerate}[leftmargin=*, nosep]
    \item We propose the \textbf{Real Deep Research (RDR)} pipeline, a systematic framework for exploring and analyzing any research area in depth.
    \item Leveraging domain expertise, we deliver high-quality survey outputs in the fields of AI and robotics, providing valuable insights for researchers and practitioners.
    \item We quantitatively evaluate the RDR pipeline and demonstrate its advantages over existing commercial large language model tools within the targeted research domains.
\end{enumerate}
\vspace{-5pt}
\section{Related Work}

\noindent
\textbf{Surveys of Foundation Models.}
In recent years, a number of survey studies have systematically reviewed foundation models across different domains~\cite{bommasani2021opportunities,Huang2022TowardsRI,Gallegos2023BiasAF,liu2023survey,zhou2024comprehensive,wang2025graph}, including natural language processing~\cite{zhao2023survey,chang2024survey}, computer vision~\cite{liu2023survey,zhang2023comprehensive}, graph learning~\cite{wang2025graph}, and robotics~\cite{zeng2023large,xu2024survey,ma2024survey,xiao2025robot}. However, these surveys require extensive manual effort and become outdated quickly due to the rapid progress of foundation models. Unlike such static surveys, our goal is to design a framework that can automatically analyze thousands of papers and provide an always up to date understanding of different research areas.

\noindent
\textbf{LLMs in Scientific Research.}
Large language models (LLMs) have been applied across various stages of scientific research~\cite{van2023ai,lu2024ai,messeri2024artificial,schmidgall2025agent}, including idea generation~\cite{wang2024scimon,baek2024researchagent}, coding~\cite{xu2022systematic,majumder2024discoverybench}, paper reviewing~\cite{liang2024can,lu2024ai}, and predicting experimental results~\cite{manning2024automated,luo2025large}. Among these stages, literature analysis plays a central role, involving tasks such as paper retrieval, clustering, and topic trend analysis. However, traditional literature search tools such as Google Scholar rely mainly on lexical matching and struggle with tasks that require deeper semantic reasoning. This has motivated researchers to leverage LLMs for literature analysis~\cite{ajith2024litsearch,li2024scilitllm,he2025pasa,shi2025spar}. For example, 
SciLitLLM~\cite{li2024scilitllm} employs supervised learning to build a specialized LLM for scientific literature understanding; PaSa~\cite{he2025pasa} uses reinforcement learning with synthetic data to train an LLM agent that can answer complex scholarly queries. Unlike prior work that focuses mainly on research question answering, our approach targets a broader and systematic understanding of entire research areas. We highlight not only semantic reasoning over large collections of papers but also automatic analysis of research trends, offering researchers a transparent and evidence-based view of the literature.

\noindent
\textbf{Knowledge Organization and Discovery.} 
It has been shown that LLMs are capable of clustering documents~\cite{viswanathan2024large,zhang2023clusterllm} and uncovering latent topics~\cite{pham2023topicgpt,li2025scitopic}. For example, Knowledge Navigator~\cite{katz2024knowledge} combines LLMs with clustering techniques to organize and structure documents for scientific literature search; SciTopic~\cite{li2025scitopic} enhances LLMs in identifying topic structures by refining document embeddings. Beyond knowledge organization, recent research~\cite{krenn2020predicting,krenn2023forecasting,gu2025forecasting} also studies the trend of high-impact research topics.
Our work introduces a novel approach by leveraging the reasoning capabilities of LLMs and the embedding representations of foundation models, which leads to more accurate and semantic knowledge organization. Built on this knowledge structure, our framework enables analysis of past and future research trends and supports inspection of connections between topics, providing valuable insights into scientific directions.
\section{Method}
\label{sec:method}

\begin{figure*}[t] % or [H], [ht], etc.
    \centering
    \includegraphics[width=1.\textwidth]{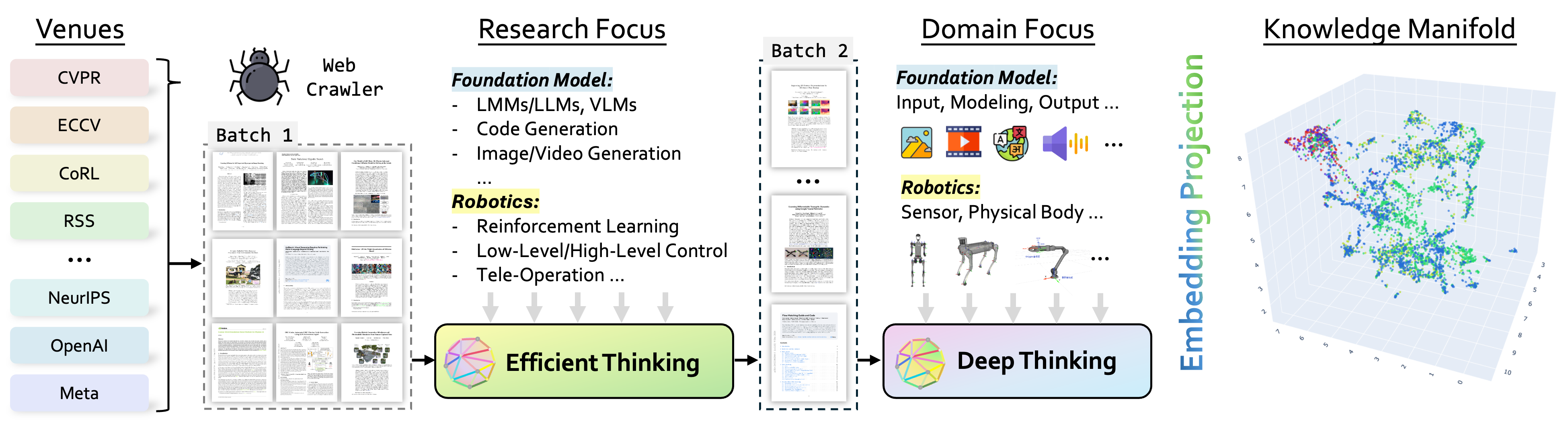}
    \vspace{-20pt}
    \caption{Pipeline of the proposed method on filtering and projecting thousands of papers to the embedding space for future analysis.}
    \label{fig:pipeline1}
    \vspace{-10pt}
\end{figure*}

In the Methods section, we focus specifically on the domains of foundation models and robotics to provide a comprehensive overview of how we conduct Real Deep Research using expert knowledge. As illustrated in Fig.~\ref{fig:pipeline1}, the embedding-based analysis pipeline consists of four main components:(1) Data Preparation (Sec.~\ref{sec:data}), (2) Content Reasoning (Sec.~\ref{sec:reasoning}), (3) Content Projection (also in Sec.~\ref{sec:reasoning}), and (4) Embedding Analysis (Sec.~\ref{sec:embedding}). This pipeline is powered by a suite of large language and multimodal models (LLMs/LMMs) for content extraction and reasoning, and is designed to be generalizable for the automated analysis of other research domains in the future. The following sections introduce each component in detail.

\subsection{Data Preparation.}
\label{sec:data}

\noindent
\textbf{Selection.}
To systematically investigate the integration of foundation models and robotics at scale, we focus on emerging trends and research priorities in both academia and industry. To capture the latest developments, we review recent publications from leading conferences in computer vision, robotics, and machine learning. Specifically, we collect papers via web crawling from top conference venues (CVPR, ECCV, ICCV, CoRL, RSS, ICRA, NeurIPS, etc.) as well as from industry research platforms (Nvidia, Meta, and OpenAI, etc.). This curated corpus comprehensively overviews the research contents in foundation models and robotics, highlighting key technical advancements, existing challenges, and future research directions. Specifically, we collect paper titles, authors, abstracts, and PDF links directly from conference and company websites. Then, we apply an area filtering process on paper titles and abstracts using an efficient LLM with a predefined set of criteria to ensure relevance to this study.

\noindent
\textbf{Area Filtering.} We define the collected paper set as $\mathbf{P}$, while it generally fall under the broad area of vision, language, machine learning, and robotics, it is not guaranteed that each paper directly aligns with the specific focus of our work, such as foundation models ($\mathbf{D}_f$) and robotics ($\mathbf{D}_r$). To address this, we introduce \textit{Area Filtering}—a step that leverages an efficient LLM with curated prompts—to identify papers relevant to our research scope. To ensure a correct filtering, we first define the scope of foundation models and robotics, clarifying technical boundaries between domains. Below are the prompts that we designed for our research focus:

\noindent%  ← move starting point 20 pt left
\colorbox{gray!10}{%
\begin{minipage}{\dimexpr\columnwidth\relax}% ← make box 20 pt wider
\footnotesize
\texttt{Foundation Model Definition: ''Research involving deep learning models (especially transformer-based) trained on large amounts of data and capable of fitting generalized factual realities. These models typically serve as versatile backbones for a variety of downstream tasks across multiple domains.''
}\\
\texttt{Key Indicators:}\\
\texttt{- Large Multimodal Models (LMM)}\\
\texttt{- Large Language Models (LLM) ...}
\end{minipage}
}

\noindent%  ← move starting point 20 pt left
\colorbox{gray!10}{%
\begin{minipage}{\dimexpr\columnwidth\relax}% ← make box 20 pt wider
\footnotesize
\texttt{Robotics Definition: ''Research involving hardware systems equipped with input sensors and mechanical kinematics capable of producing joint movements. These systems are controlled by learning-based algorithms that facilitate automatic or robust mappings from sensory inputs to actuator outputs.''
}\\
\texttt{Key Indicators:}\\
\texttt{- Reinforcement Learning in robotic contexts}\\
\texttt{- Imitation Learning for physical systems ...}
\end{minipage}
}

After filtering using an efficient LLM, the resulting set of papers (\(\mathbf{P}'\)) belongs to either the foundation model domain (\(\mathbf{D}_f\)), the robotics domain (\(\mathbf{D}_r\)), or both. Formally we write \(\mathbf{P}' = \{ p \mid p \in \mathbf{D}_f \cup \mathbf{D}_r \}\).

\vspace{-5pt}
\subsection{Content Reasoning.}
\label{sec:reasoning}
Given the filtered papers $\mathbf{P}'$ in the domains of foundation models and robotics, an in-depth analysis is required to narrow the position of each paper. Guided by domain experts in foundation models and robotics, we define perspectives that align with established domain structures, emerging trends, and evolving knowledge. Beyond predefined perspectives, our pipeline supports future user-defined perspectives, allowing adaptation to new research questions. In the following paragraphs, we will depict the general structure, trends, and knowledge of the foundation model and robotics in preparation for analyzing the research works under $\mathbf{P}'$.

\begin{figure*}[t] % or [H], [ht], etc.
    \centering
    \includegraphics[width=1.\textwidth]{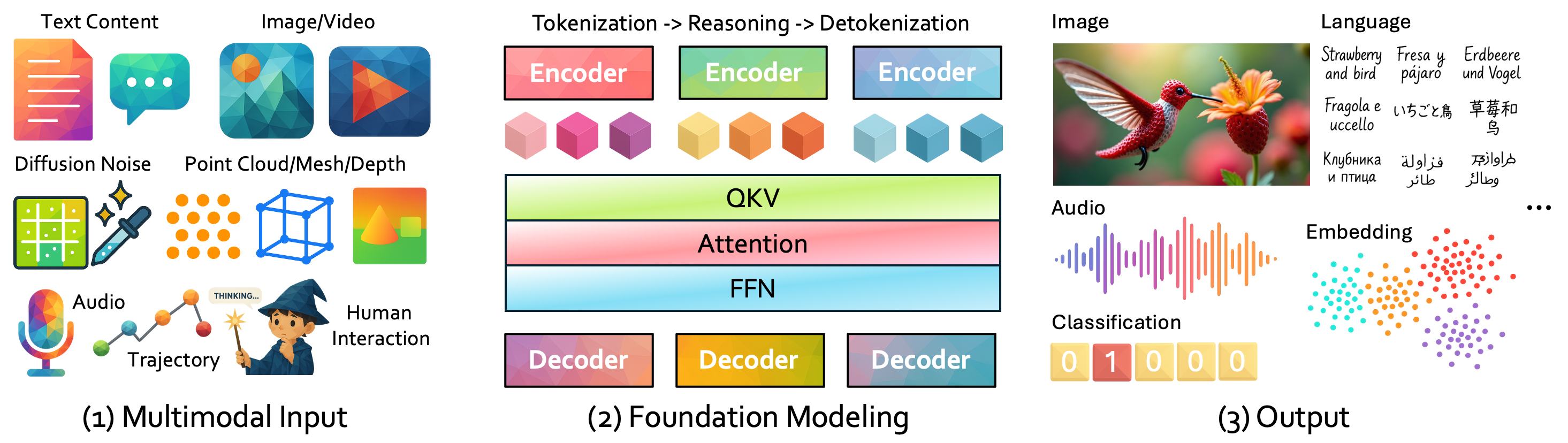}
    \vspace{-20pt}
    \caption{Perspective analysis of foundation model research, which primarily includes (1) Input, (2) Modeling, (3) Output, etc., shown in the figure.}
    \label{fig:foundation}
    \vspace{-10pt}
\end{figure*}

\noindent
\textbf{Foundation Model.}
A foundation model’s development are systematically analyzed through five fundamental perspectives in this work: Input ($\mathbf{I}$), Modeling ($\mathbf{M}$), Output ($\mathbf{O}$), Objective ($\mathbf{W}$), and Learning Recipe ($\mathbf{R}$). We have shown some main perspectives examples in Fig.~\ref{fig:foundation}.  This structured representation facilitates a comprehensive analysis of the foundation model. Below is the formal writing for the procedure:
\vspace{-2pt}
\[
\mathcal{D}_{f}^{P'} = \bigcup_{p \in \mathbf{P}'} F(p), \quad F(p) = \text{LLM}(p \mid \mathbf{I}, \mathbf{M}, \mathbf{O}, \mathbf{W}, \mathbf{R}),
\vspace{-5pt}
\]
where LLM represents the large multimodal model, and $\mathcal{D}_{f}^{P'}$ denotes the perspective projection of the given papers in $\mathbf{P}'$, focused on foundation model research. In the following paragraphs, we provide a formal definition of each perspective:

\noindent
\textit{Input ($\mathbf{I}$). The input processing for a foundation model generally involves raw data and a tokenization procedure. Standard input sources include images, videos, audio, LiDAR, etc., with tokenization performed through transformations and neural networks.}

\noindent
\textit{Modeling ($\mathbf{M}$). With input settled for a foundation model, the modeling part is responsible for extracting critical knowledge from the input, reasoning, and decoding to the output space. It is the critical procedure to transfer input knowledge to output.}

\noindent
\textit{Output ($\mathbf{O}$). The task determines the decoding space according to the input and modeling, this is the final step to decode the latent representation to the output used for loss computation or the final interaction.}

\noindent
\textit{Objective ($\mathbf{W}$). To fit a foundation model with the corresponding input, and output, the given model architecture is constrained by the learning objective, this fits the model distribution in alignment with the transformation given the task(s).}

\noindent
\textit{Recipe ($\mathbf{R}$). The recipe is used as the cookbook on how to tune the model weight with input, output, and objective. It controls the training stage, convergence speed, and updated parameters.}

\begin{figure*}[t] % or [H], [ht], etc.
    \centering
    \includegraphics[width=1.\textwidth]{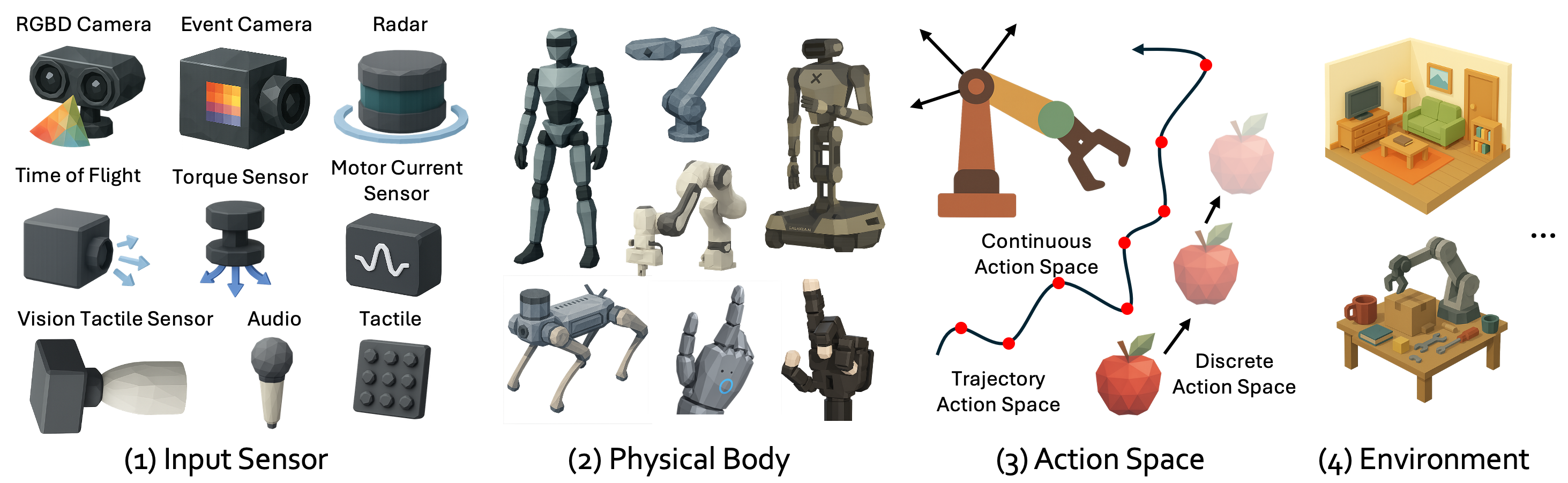}
    \vspace{-20pt}
    \caption{Perspective analysis of robotics research, which primarily includes (1) Input, (2) Modeling, (3) Output, etc., shown in the figure.}
    \label{fig:robot}
    \vspace{-6pt}
\end{figure*}

\noindent
\textbf{Robotics.} 
For research work in robotics, the core perspective shifts to emphasize hardware and interaction within real-world environments. We define five key perspectives to map each paper within the broader landscape of robotic applications: Input Sensor ($\mathbf{S}$), Physical Body ($\mathbf{B}$), Joint Output ($\mathbf{J}$), Action Space ($\mathbf{A}$), and Environment ($\mathbf{E}$). An example of core perspectives is illustrated in Fig.~\ref{fig:robot}.
These perspectives collectively define how robots perceive, act, and interact within the physical world. The procedure could be formally written as:
\[
\vspace{-4pt}
\mathcal{D}_{r}^{P'} = \bigcup_{p \in \mathbf{P}'} F(p), \quad F(p) = \text{LMM}(p \mid \mathbf{S}, \mathbf{B}, \mathbf{J}, \mathbf{A}, \mathbf{E}),
\vspace{-4pt}
\]
where $\mathcal{D}_{r}^{P'}$ represents the perspective projection of the given papers in $\mathbf{P}'$, providing a structured framework for analyzing robotics research. We show the concrete definition in the following:

\noindent
\textit{Input Sensor ($\mathbf{S}$). Input sensors are hardware devices that measure physical quantities or environmental conditions and convert them into digital signals that can be processed by the robot's control system. They serve as the robot's interface with its environment.}
\vspace{-1pt}

\noindent
\textit{Physical Body ($\mathbf{P}$). A physical body in robotics refers to the mechanical structure and architecture that enables physical interaction with the environment. This physical manifestation determines how motor commands translate into real-world forces, movements, and environmental manipulations.}
\vspace{-1pt}

\noindent
\textit{Action Space ($\mathbf{A}$). The action space is the set of all permissible actions a robot can select in a given context, ranging from low-level joint commands to high-level behaviors (e.g., “walk” or “grasp”). Each chosen action is ultimately executed as a joint output, bridging decision-making to physical movement.}
\vspace{-1pt}

\noindent
\textit{Joint Output ($\mathbf{J}$). Joint output refers to the physical movement or configuration of a robot's joints resulting from executed motor commands. It translates control signals (e.g., torque or velocity) into mechanical motion, allowing the robot to directly interact with and manipulate its environment.}
\vspace{-1pt}

\noindent
\textit{Environment ($\mathbf{E}$). The environment encompasses the physical space where a robot operates, characterized by its spatial layout, structural features, and contextual elements (e.g., furniture, tools, obstacles) that shape the task-specific challenges and opportunities the robot encounters.}

Given the predefined perspective definition, we use the following prompt to extract each perspective from the paper:

\noindent%  ← move starting point 20 pt left
\colorbox{gray!10}{%
\begin{minipage}{\dimexpr\columnwidth\relax}% ← make box 20 pt wider
\footnotesize
\texttt{Can you analyze the paper contents according to the following perspectives: (1) Definition 1, (2) Definition 2, (3) Definition 3, ...
}\\
\texttt{After analysis, please identify each of the perspectives in the paper, and return the answer in the following format: \{"perspective 1": plain text, "perspective 2": plain text, "perspective 3": plain text, ...\}}
\end{minipage}
}

\subsection{Content Projection.}
\label{sec:projection}

Given the extracted contents from research papers guided by our defined perspective, we aim to project natural language descriptions into an informative latent space. This projection enables large-scale analysis of current research in foundation models and robotics while revealing potential future research directions. Motivated by recent advancements in large language model-based embedding models, we \textbf{employ a pre-trained embedding foundation model} $\mathbf{G}$ to project $\mathcal{D}_{r}^{P'}$ (processed robotics papers' content) and $\mathcal{D}_{f}^{P'}$ (processed foundation model papers' content) from natural language space into a more abstract embedding space. The embedding model maps text into a high-dimensional vector space where semantically similar concepts occupy proximate regions.

We formally define this projection procedure as follows: For any text snippet $x \in \mathcal{D}$, its embedding is computed as: \(v_x = \mathbf{G}(x) \in \mathbb{R}^d\). Our core assumption is that by projecting paper contents through this perspective-aware embedding process and analyzing them in the high-dimensional manifold, we can uncover meaningful patterns, research trends, and potential gaps in the literature through systematic visualization and clustering analysis.

\vspace{-5pt}
\subsection{Embedding Analysis.}
\label{sec:embedding}
The goal of embedding analysis is to structure the understanding of previously extracted embeddings. The pipeline for embedding analysis contains three components: (1) Clustering on the extracted embeddings and analyzing the main concept from each cluster. (2) Structured the concept for each cluster to formulate an informative table. (3) Given the structured understanding, we trace back to the reference papers.

\textbf{Clustering for Embeddings.} We first embed every paper to obtain its vector representation $V$ and partition the corpus into $k$ clusters. From each cluster, we then draw a random sample of 50 papers and feed their text to a reasoning-based model with the prompt:

\noindent%  ← move starting point 20 pt left
\colorbox{gray!10}{%
\begin{minipage}{\dimexpr\columnwidth\relax}% ← make box 20 pt wider
\footnotesize
\texttt{Can you summarize the following contents into three distinct keywords:
    Here is one example output:"keyphrase1, keyphrase2, keyphrase3".
    The output should be short and precise, with a single output for all papers.
}
\end{minipage}
}

The model returns three compact key phrases that capture the cluster’s core theme, giving every paper both a cluster label and an interpretable set of keywords for subsequent analysis.

\textbf{Structuring for Thoughts.}
With clustered embeddings and their associated topic keywords in place, the next step is to generate a structured survey for the given research area. To accomplish this, we leverage the o3 language model, using the clustered keywords as prompts to guide the formulation of the final survey content. Incorporating the clustering results into the prompt ensures that the generated text remains grounded in the actual structure of the research landscape, enhancing both coherence and relevance. We use the following prompt to produce the final output:

\noindent%  ← move starting point 20 pt left
\colorbox{gray!10}{%
\begin{minipage}{\dimexpr\columnwidth\relax}% ← make box 20 pt wider
\footnotesize
\texttt{
Those are summarized keywords for a number of science papers clustered by abstract contents, 
however they are ambiguous, contents may overlap between clusters, 
can you summarize the information in a more structured way for audience with the following criteria: ...
}
\end{minipage}
}
\section{Analysis}
In this section, we conduct a comprehensive qualitative analysis of the conclusions drawn in this work from the following perspectives: (1) Embedding analysis for general research areas. (2) Embedding analysis within specific perspective. (3) Trend analysis of research focus over time. (4) Knowledge graph exploration across different research areas. (5) Retrieval examples based on embeddings. This pipeline will enable a researcher to dive into any research area, identify what to explore, and determine the specific papers to focus on.

\noindent
\textbf{Embedding Analysis - General.} 
The output of the embedding analysis is a comprehensive survey tailored to the featured research domain. This survey is organized into major categories and sub-categories, each detailing the specific topics covered. Rather than generating the survey content via LLM, we leverage the clustering results from the embedding analysis to guide its structure and scope. Additionally, for each sub-topic, we include the most relevant citations to provide readers with direct references for further exploration. We have provide the full survey for Foundation Model, Robotics, Computer Vision, Natural language processing, and machine learning in Appendix.

\begin{table}[ht]
\centering\scriptsize
\begin{tabularx}{\textwidth}{
  @{} >{\raggedright\arraybackslash}p{0.02\linewidth}
      >{\raggedright\arraybackslash}p{0.23\linewidth}
      >{\raggedright\arraybackslash}X
      >{\raggedright\arraybackslash}p{0.34\linewidth}
      >{\centering\arraybackslash}p{0.06\linewidth}@{}}
\toprule
\textbf{Cat.} & \textbf{Sub-category} & \textbf{What is covered} &
\textbf{Typical examples} & \textbf{\scriptsize Cluster} \\
\midrule

%=========== 1. Perception & Mapping ===========
\rowcolor{CategoryBG}
\multicolumn{5}{@{}l}{\bfseries 1.\;Perception \& Mapping \cite{AbbasSadat2024Perceive,JuexiaoZhang2024Multiview,WenzhaoZheng2024Occworld,MedhiniNarasimhan2024Seeing,Harwath2024Jointly}}\\
& 1.1 Multimodal sensor fusion \cite{Liang2024Deep,Yin2024Fusion,TimBroedermann2024Muses,Zimmer2024Tumtraf,TsunHsuanWang2024V}
& Fuse heterogeneous sensors for richer scene understanding
& LiDAR–Camera Fusion; Radar–Camera Fusion; V2X Cooperative Perception ...
& \scriptsize 0,6,7,8,14,16 \\

\rowcolor{AltGray}
& 1.2 3D reconstruct/occupancy \cite{WenzhaoZheng2024Occworld,Huang2024Selfocc,HaisongLiu2024Fully,WeiyangLiu2024Structural,Peng2024Learning}
& Build dense or sparse geometric maps for localisation
& 3-D SLAM \& Reconstruction; 3-D Occupancy; Efficient 3-D Representation
& \scriptsize 0,8,16 \\

& 1.3 BEV / top-view mapping \cite{Zhao2024Improving,Yin2024Fusion,Chambon2024Pointbev,KaiJiang2024Da,ZhiqiLi2024Bevformer}
& Bird’s-eye or top-down representations for planning
& BEV Perception; V2X Collaborative Perception
& \scriptsize 0,14,16 \\

\rowcolor{GREEN}
\multicolumn{5}{@{}l}{\textbf{... ...}}\\

\bottomrule
\end{tabularx}
\label{tab:ea_general}
\end{table}

\noindent
\textbf{Embedding Analysis - Perspective.} 
After establishing a clear overview of the domain, we analyze it through a targeted \emph{perspective} to expose structure and problem formulations. As introduced in the Methods section, our perspective analysis uses embedding-based clustering to organize works along a chosen axis. In this study we focus on foundation models and robotics, examining how each community formulates its problems. Below we illustrate the robotics case from the viewpoint of \emph{action space}. This perspective-guided embedding analysis yields a deeper understanding of the domain and a high-level map of how researchers approach and solve its problems. We also provide the full perspective for foundation model and robotics in Appendix.

\begin{table}[ht]
\centering\scriptsize
\begin{tabularx}{\textwidth}{
  @{} >{\raggedright\arraybackslash}p{0.1\linewidth}
      >{\raggedright\arraybackslash}p{0.18\linewidth}
      >{\raggedright\arraybackslash}X
      >{\raggedright\arraybackslash}p{0.29\linewidth}
      >{\centering\arraybackslash}p{0.10\linewidth}@{}}
\toprule
\textbf{Category} & \textbf{Sub-category} & \textbf{What is covered} &
\textbf{Typical examples} & \textbf{\scriptsize Cluster} \\
\midrule

%=========== 1. Continuous Low-Level Actuation ===========
\rowcolor{CategoryBG}
\multicolumn{5}{@{}l}{\bfseries 1.\;Continuous Low-Level Actuation}\\
& 1.1 Joint-space commands
& Direct numerical inputs to individual joints or actuators, bounded by hardware limits.
& joint torques/positions/velocities; high-dimensional joint commands; bounded control inputs; finger-joint configs; parametrised joint trajectories
& \scriptsize 0, 4, 6, 10, 12, 14, 18 \\

\rowcolor{AltGray}
& 1.2 Vehicle / body dynamics commands
& Low-level controls that change a mobile base, ground-vehicle or aerial body state.
& steering angle; throttle / acceleration; braking; linear \& angular velocity; body-rate thrust; speed/direction for locomotion; lane-keeping
& \scriptsize 0, 1, 7, 10, 12, 13, 15 \\

\rowcolor{GREEN}
\multicolumn{5}{@{}l}{\textbf{... ...}}\\
\bottomrule
\end{tabularx}
\label{tab:ea_perspective}
\end{table}

\noindent
\textbf{Trend Analysis.}
Once we understand each domain and its key sub-perspectives, the next step is to assess topic momentum. Our trend analysis highlights which areas are accelerating and which have been thoroughly explored in recent years, giving a practical starting point when entering a new field. In robotics (see figure), the trajectories suggest that teleoperation, dexterous manipulation, and low-cost open-source robotics are currently rising, while traditional reinforcement learning and skill-based manipulation appear comparatively mature or show slowing activity. This will guide the researchers to smoothly enter a new field. We provide the full trend analysis in Appendix.~\ref{sec:trend} for Computer Vision, NLP, Robotics, and Machine Learning.

\begin{figure*}[htbp]
    \centering
    \tiny
    \renewcommand{\arraystretch}{1.4}
    \setlength{\tabcolsep}{4pt}
    \begin{tabular}{|p{0.30\textwidth}|p{0.30\textwidth}|p{0.30\textwidth}|}
    \hline
    % --------------------------- Row 1 -------------------------------
    \begin{minipage}[t]{\linewidth}\centering
        \includegraphics[width=0.9\linewidth]{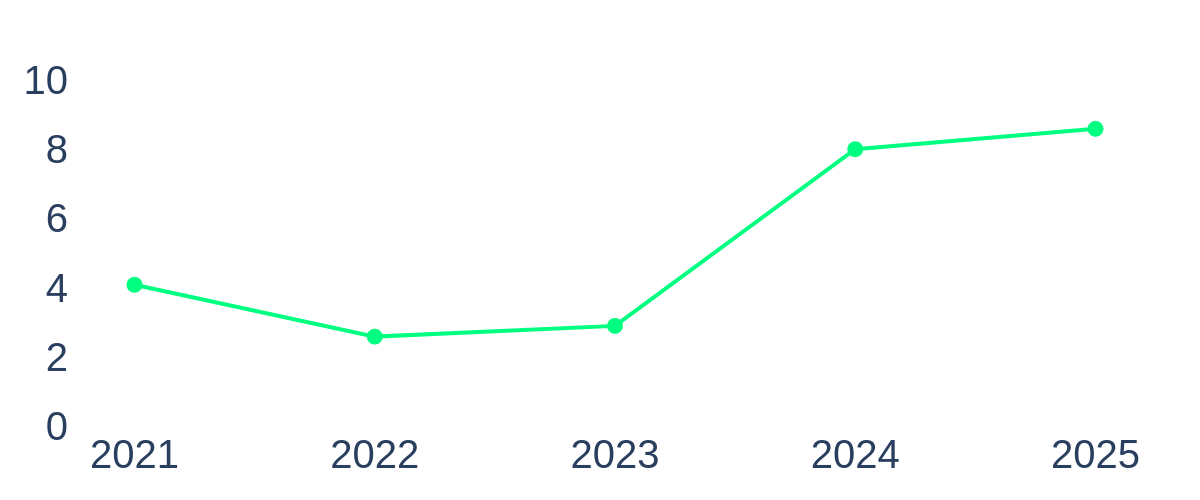}\par
        1.\;Legged Locomotion, Reinforcement Learning, Sim-to-Real Transfer
    \end{minipage} &
    \begin{minipage}[t]{\linewidth}\centering
        \includegraphics[width=0.9\linewidth]{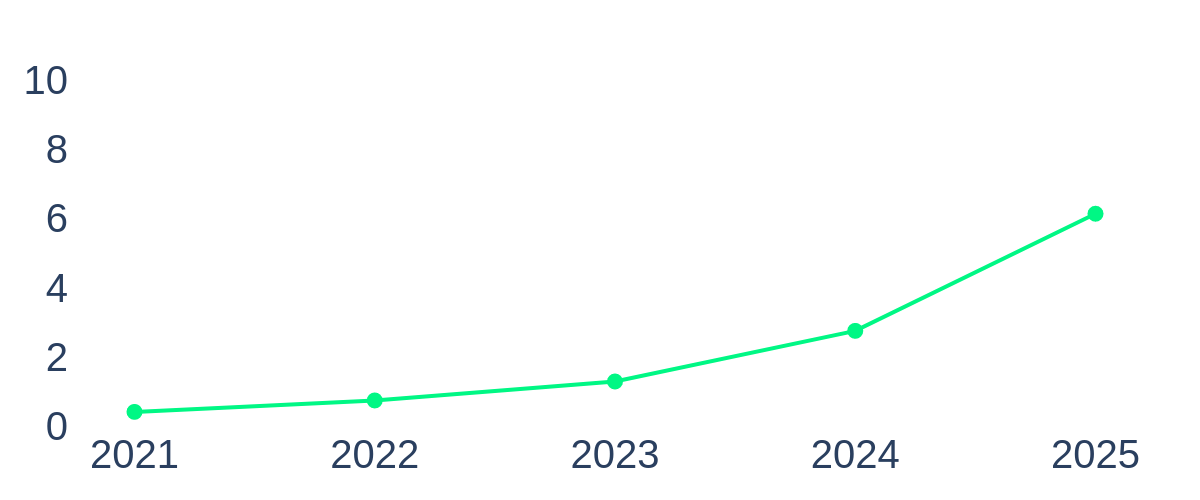}\par
        2.\;Teleoperation, Dexterous Manipulation, Low-Cost Open-Source Robotics
    \end{minipage} &
    \begin{minipage}[t]{\linewidth}\centering
        \includegraphics[width=0.9\linewidth]{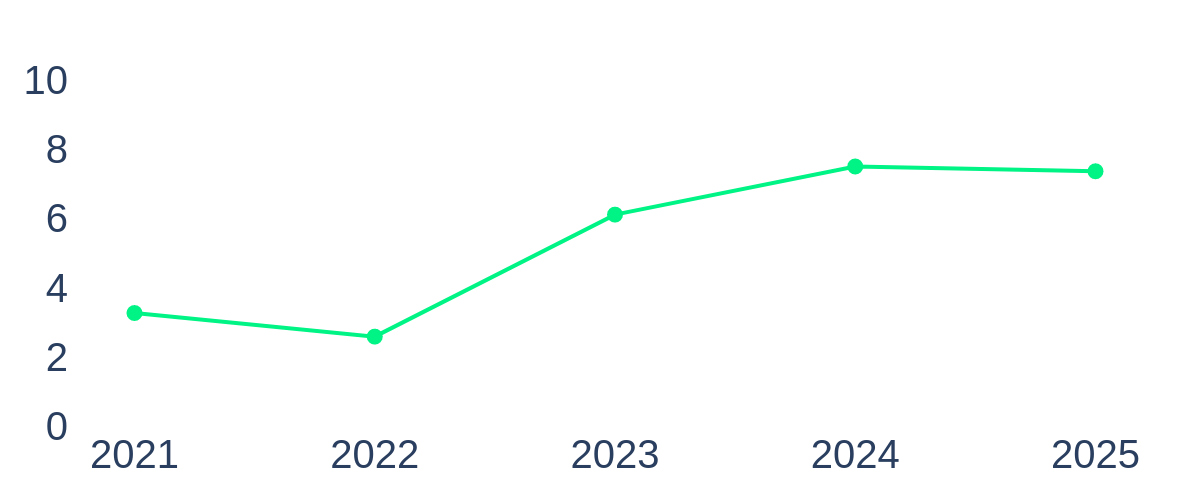}\par
        3.\;Language-Conditioned Manipulation, Vision-Language-Action Models, 3D Scene Grounding
    \end{minipage}\\ \hline
    \begin{minipage}[t]{\linewidth}\centering
    \textbf{... ...}
    \end{minipage} &
    \begin{minipage}[t]{\linewidth}\centering
    \textbf{... ...}
    \end{minipage} &
    \begin{minipage}[t]{\linewidth}\centering
    \textbf{... ...}
    \end{minipage}\\ \hline
    % --------------------------- Row 10 ------------------------------
    \begin{minipage}[t]{\linewidth}\centering
        \includegraphics[width=0.9\linewidth]{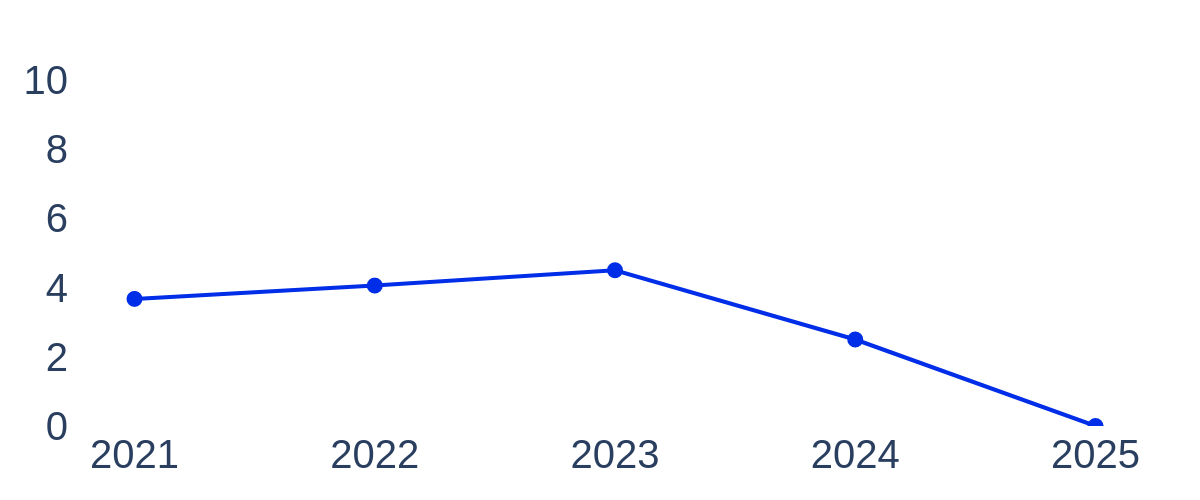}\par
        28.\;Offline Reinforcement Learning, Robotic Skill Learning, Continual Adaptation
    \end{minipage} &
    \begin{minipage}[t]{\linewidth}\centering
        \includegraphics[width=0.9\linewidth]{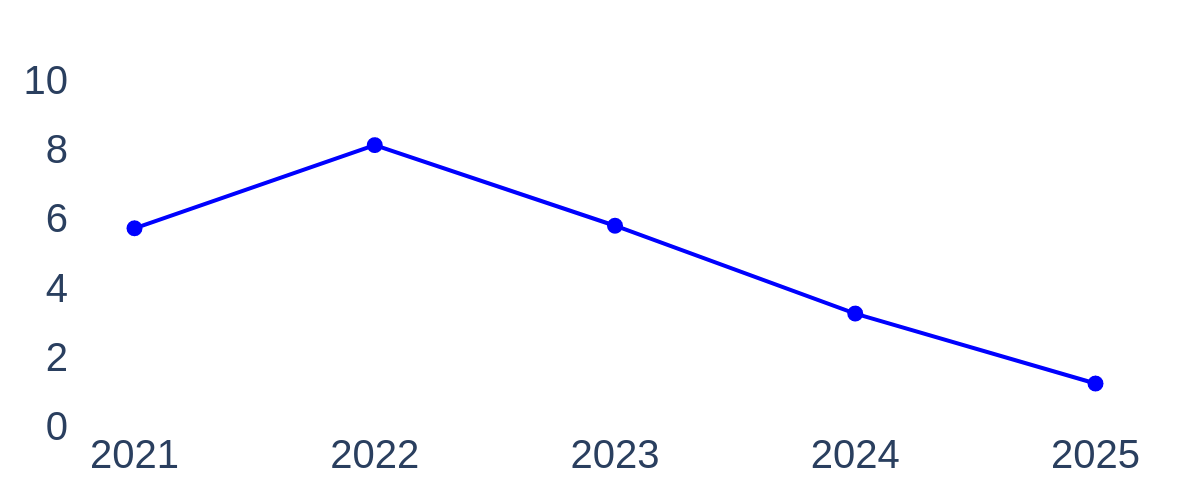}\par
        29.\;Trajectory Prediction, Safety-Critical Scenario Generation, Autonomous Driving Simulation
    \end{minipage} &
    \begin{minipage}[t]{\linewidth}\centering
        \includegraphics[width=0.9\linewidth]{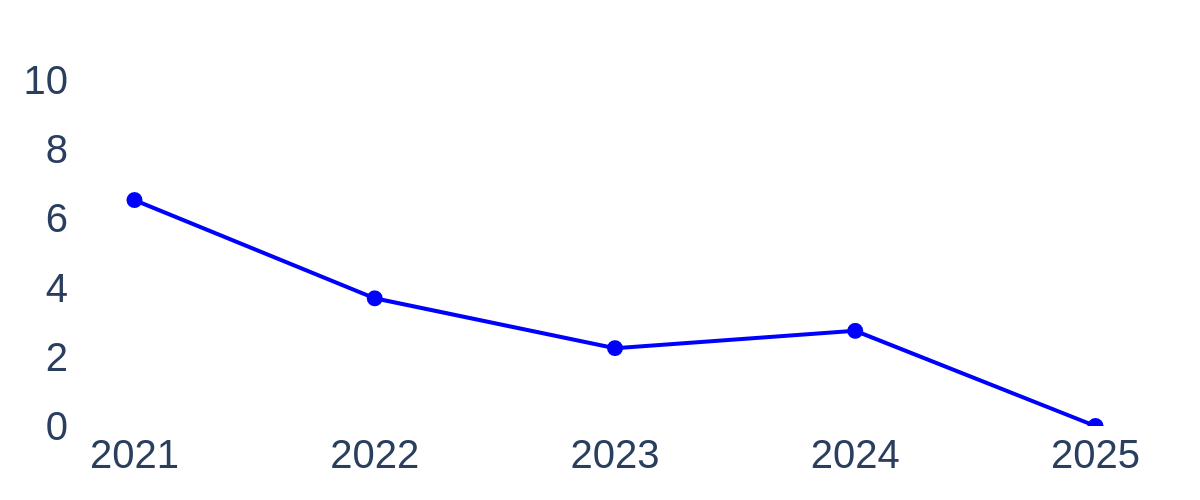}\par
        30.\;Robotic Reinforcement Learning, Skill-Based Manipulation, Sample-Efficient Learning
    \end{minipage}\\ \hline
    \end{tabular}
    \label{trend}
\end{figure*}

\noindent
\textbf{Knowledge Graph.}
Beyond identifying trending topics within individual research areas, an equally important direction for discovery lies in uncovering cross-domain themes—topics that span multiple fields and have the potential to catalyze interdisciplinary breakthroughs. In this work, we analyze the intersections among four major domains: computer vision, natural language processing, machine learning, and robotics. By examining these intersections, we aim to highlight not only where collaboration already exists but also where it could be further cultivated.

As shown in Figure below, the left side of the figure presents a Cross-Domain Topology Graph, where each color corresponds to a specific research domain. Each node (represented as a sphere) signifies a distinct topic cluster derived from our embedding-based analysis, and edges between nodes indicate semantic or topical relationships—especially those that cross domain boundaries. Nodes located at the periphery, with few or no connecting edges, represent domain-specific topics that remain largely isolated from other fields. In contrast, the densely connected regions at the center of the graph reflect genuinely cross-domain topics, where ideas, methods, or applications from multiple fields converge. This view empowers researchers to discover promising frontiers for collaboration, encourages rethinking isolated problems through new lenses, and supports a more forward-looking approach to scientific inquiry.

\begin{figure*}[htbp] % or [H], [ht], etc.
    \vspace{-5pt}
    \centering
    \includegraphics[width=1.\textwidth]{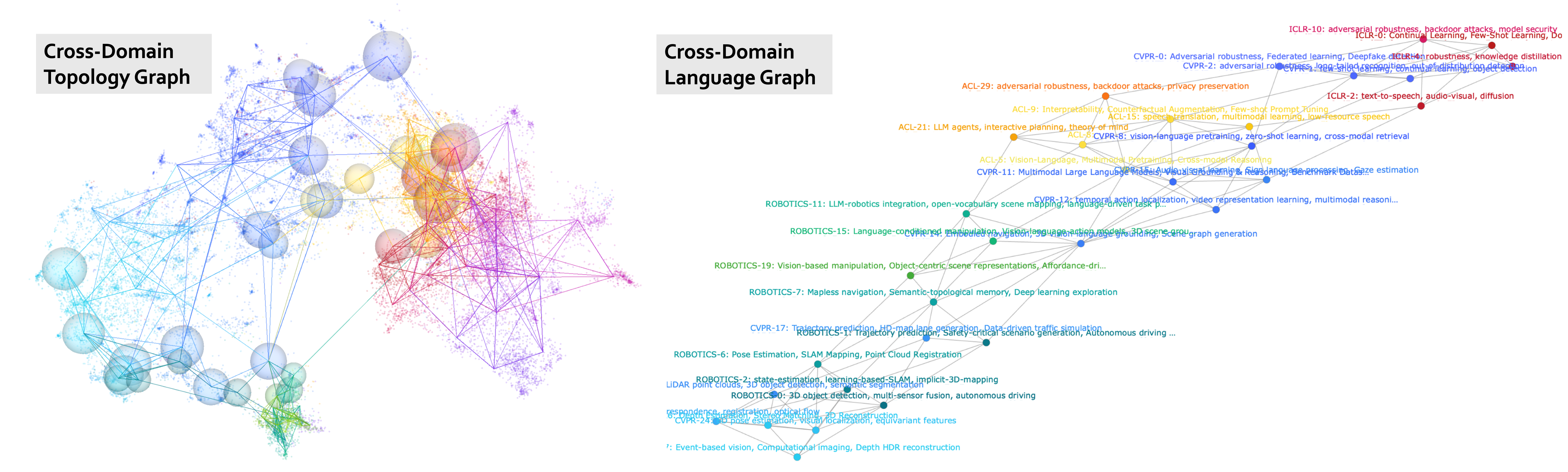}
    \vspace{-15pt}
    \label{fig:cross_domain}
\end{figure*}

\noindent
\textbf{Retrieval Examples.}
Once a target research topic is identified, the next step is to pinpoint concrete entry points. We do this by leveraging the conference-level embeddings inferred earlier to run semantic searches and retrieve the most relevant literature. For example, after surveying robotics, we focus on dexterous manipulation and query the embedding index to surface closely related papers across venues. As shown in the table below, the returned papers align tightly with the query and exhibit meaningful community impact, as reflected by their venues, years, and citation counts.

\begin{table}[ht]
\vspace{-5pt}
\centering
\small
\begin{tabularx}{\textwidth}{
  @{}
  >{\raggedright\arraybackslash}X
  >{\centering\arraybackslash}c
  >{\centering\arraybackslash}c
  >{\centering\arraybackslash}c@{}
}
\toprule
\textbf{Paper} & \textbf{Year} & \textbf{Venue} & \textbf{Citations} \\
\midrule

\rowcolor{CategoryBG}
\multicolumn{4}{@{}p{\dimexpr\textwidth-2\tabcolsep}}{\bfseries Query: dexterous manipulation generated data in 3D simulation and evaluated in real world.} \\
Evaluating Real-World Robot Manipulation Policies in Simulation & 2024 & CoRL24 & 127 \\
\rowcolor{AltGray}
Lessons from Learning to Spin ``Pens'' & 2024 & CoRL24 & 29 \\
General In-hand Object Rotation with Vision and Touch & 2023 & CoRL23 & 134 \\
\rowcolor{AltGray}
Twisting Lids Off with Two Hands & 2024 & CoRL24 & 13 \\
DexterityGen: Foundation Controller for Unprecedented Dexterity & 2025 & RSS25 & 16 \\
\bottomrule
\end{tabularx}
\label{tab:retrieved_papers_with_venue}
\vspace{-10pt}
\end{table}

\vspace{-5pt}
\section{Experiment}
\vspace{-5pt}
We have presented a comprehensive qualitative analysis demonstrating how Real Deep Research supports deep dives into a chosen research focus. This section now details the dataset curated for our study and the implementation specifics required to realize the system. We then provide quantitative evaluations—both by benchmarking our survey against commercial research tools and by validating the effectiveness of the embeddings that underpin our approach.

\begin{wraptable}{r}{0.5\textwidth} % 'r' aligns to the right, '0.5\textwidth' sets width
    \vspace{-25pt}
    \centering
    \begin{tabular}{lllr}
    \toprule
    \textbf{Venue} & \textbf{Year} & \textbf{Area} & \textbf{Total} \\ 
    \midrule
    CVPR & 21-25 & \textit{Computer Vision} & 11668 \\
    CoRL & 21-24 & \textit{Robotics} & 815 \\
    RSS & 21-25 & \textit{Robotics} & 575 \\
    ICLR & 21-25 & \textit{Machine Learning} & 9549 \\
    ACL & 21-25 & \textit{NLP} & 4556 \\
    NeurIPS & 2024 & \textit{Machine Learning} & 4240 \\
    ECCV & 2024 & \textit{Computer Vision} & 6166 \\
    \bottomrule
    \end{tabular}
    \vspace{-5pt}
    \caption{\textbf{Paper Distribution Analysis} across different venues, showing the total number of papers.}
    \label{tab:paper_distribution}    
    \vspace{-20pt}
\end{wraptable}

\noindent
\textbf{Dataset.}
We curate our dataset from a collection of publicly available, high-impact conference venues, focusing on those central to the fields of artificial intelligence and robotics. As shown in Table~\ref{tab:paper_distribution}, the dataset includes papers from venues such as CVPR, ECCV (Computer Vision), ICLR, NeurIPS (Machine Learning), ACL (Natural Language Processing), and CoRL and RSS (Robotics). This selection allows us to capture a broad yet targeted view of the AI and robotics research landscape from 2021 through 2025. To align with our focus on foundation models and robotics, we apply an additional filtering step across all venues. Specifically, we identify and extract 4,424 papers related to foundation models and 1,186 papers focused on robotics, both from the year 2024 onward. This subset enables us to track the most recent developments in these fast-moving areas with higher resolution.

\begin{table}[t]
\centering
\resizebox{\textwidth}{!}{
\begin{tabular}{@{}l|c|*{10}{c}@{}}
\toprule
& & \multicolumn{4}{c}{\textbf{General}} & \multicolumn{3}{c}{\textbf{Foundation Model}} & \multicolumn{3}{c}{\textbf{Robotics}} \\
\cmidrule(lr){3-6} \cmidrule(lr){7-9} \cmidrule(lr){10-12}
\textbf{Model} & \textbf{Rank} & CV & NLP & ML & Robotics & Input & Modeling & Output & Sensor & Body & Action \\
\midrule
GPT5             & 4.80 & 10.00 & 17.39 & 45.45 & 71.43 & 44.44 & 10.00 & 21.05 & 22.73 & 34.78 & 69.57  \\[0.5ex]
GPT5-Thinking    & 2.75 & \textbf{82.61} & 59.09 & 47.83 & 66.67 & 55.00 & \textbf{90.91} & 50.00 & 88.46 & 42.86 & 32.00  \\[0.5ex]
GPT5-Research    & 4.00 & 42.11 & 50.00 & 72.73 & 63.64 & 21.05 & 35.00 & 50.00 & 0.00 & 40.91 & 52.63  \\[0.5ex]
Gemini           & 4.80 & 35.00 & 40.00 & 15.38 & 0.00 & 13.64 & 54.17 & 45.83 & 31.25 & 45.00 & 26.32  \\[0.5ex]
Gemini-Thinking  & 3.35 & 63.64 & 50.00 & 56.25 & 37.50 & 65.22 & 45.45 & 41.67 & 55.56 & 56.52 & 34.78  \\[0.5ex]
\rowcolor{orange!10}
RDR (Ours)       & \textbf{1.30} & 58.33 & \textbf{89.47} & \textbf{73.68} & \textbf{77.78} & \textbf{88.46} & 60.00 & \textbf{94.74} & \textbf{91.30} & \textbf{84.21} & \textbf{89.47}  \\[0.5ex]
\bottomrule
\end{tabular}
}
\vspace{-8pt}
\caption{Survey Quality Evaluation among RDR and commercial based methods. We evaluate the pairwise winning rate for each domain and perspective.}
\vspace{-13pt}
\label{tab:model_comparison}
\end{table}

\noindent
\textbf{Implementation Details.}
We do not train any new networks in this work for generating the embedding or survey; instead, we rely on off-the-shelf models. For straightforward tasks—such as classifying research areas—we use the Doubao language model. For reasoning-intensive tasks and complex summarization, we employ the o3 model to achieve stronger performance. To extract text embeddings, we use nvidia/NV-Embed-v2.

\noindent
\textbf{Survey Quality.}
As demonstrated in Sec.~\ref{sec:method}, our analysis of a research area begins with a broad survey of the existing literature. The analysis pipeline we propose is designed to significantly reduce model hallucination and produce a comprehensive, high-quality survey for a given research direction.

To evaluate the accuracy and quality of the generated surveys, we conducted a user study involving experienced researchers with domain expertise in robotics and foundation models. As a baseline, we prompted a commercial large language model using the following instruction:
\textit{“Act as an expert research analyst. Your task is to create a structured map of the research landscape for a given academic or industrial field. The output must be a single, valid JSON object that categorizes the field into its primary research areas and specific sub-topics. For the research area 'foundation model,' can you summarize the input perspective with the following definition: The input processing for a foundation model generally involves raw data and a tokenization procedure ...”}

To assess the quality of the generated surveys, we adopted a pairwise comparison methodology rather than asking evaluators to select a single best output. For each comparison, domain experts were presented with two survey outputs and asked to determine which one demonstrated superior quality and accuracy. This evaluation setup helps reduce cognitive load and bias, making the assessment more reliable by avoiding the need for evaluators to recall or rank multiple outputs simultaneously. In total, we collected 8 evaluation entries, each with 80 pairwise comparisons. To quantify performance, we computed the winning rate of each method within its respective domain.

As shown in Tab.~\ref{tab:model_comparison}, our method, Real Deep Research (RDR), achieves the highest overall performance with an average rank of 1.30, outperforming all baselines. RDR leads in key domains such as NLP (89.47), robotics (77.78), and foundation model output (94.74), and also shows strong performance in robotics subfields like sensor (91.30) and action (89.47). While GPT5-Thinking slightly outperforms in CV (82.61) and foundation model modeling (90.91), RDR consistently ranks at or near the top across nearly all categories.

\begin{wraptable}{r}{0.5\columnwidth}
\vspace{-5pt}
\centering
\label{tab:clustering_performance_wrap}
\resizebox{\linewidth}{!}{
\begin{tabular}{l ccc ccc}
\toprule
\multirow{2}{*}{\textbf{Model}} & \multicolumn{3}{c}{\textbf{AG News}} & \multicolumn{3}{c}{\textbf{20 News Groups}} \\
\cmidrule(lr){2-4} \cmidrule(lr){5-7}
& ACC($\uparrow$) & NMI($\uparrow$) & ARI($\uparrow$) & ACC($\uparrow$) & NMI($\uparrow$) & ARI($\uparrow$) \\
\midrule
LDA & 74.05 & 47.17 & 49.01 & 29.05 & 31.63 & 13.34 \\
NMF & 34.05 & 4.59 & 2.13 & 12.42 & 12.86 & 0.48 \\
ProdLDA & 80.93 & 56.51 & 60.91 & 37.42 & 45.67 & 23.89 \\
DecTM & 55.63 & 40.04 & 36.17 & 36.57 & 46.18 & 22.90 \\
ETM & 26.14 & 0.00 & 0.00 & 5.35 & 0.10 & 0.00 \\
NSTM & 26.14 & 0.01 & 0.00 & 16.92 & 17.02 & 2.34 \\
TSCTM & 79.63 & 53.91 & 55.89 & 40.60 & 44.06 & 15.71 \\
ECRTM & 78.69 & 54.05 & 54.88 & 25.70 & 31.00 & 12.26 \\
Bertopic & 35.93 & 12.88 & 7.03 & 29.78 & 28.57 & 11.58 \\
FASTopic & 83.48 & 59.10 & 62.48 & 51.65 & 56.32 & 39.49 \\
\textcolor{gray}{SciTopic*} & \textcolor{gray}{85.29} & \textcolor{gray}{61.96} & \textcolor{gray}{65.94} & \textcolor{gray}{70.88} & \textcolor{gray}{68.32} & \textcolor{gray}{55.71} \\
\rowcolor{orange!10}
RDR (Ours) & 84.86 & 61.66 & 65.24 & 52.91 & 56.57 & 39.96 \\
\bottomrule
\end{tabular}
}
\caption{Unsupervised Clustering performance. * indicate using more labels.}
\vspace{-10pt}
\end{wraptable}

\noindent
\textbf{Embedding Quality.}
Because much of our analysis relies on high-quality embeddings, we evaluate their effectiveness using a simple linear probe trained on top of frozen representations—an approach that best reflects the intrinsic utility of the embeddings themselves. We follow the experimental protocol introduced in SciTopic~\cite{li2025scitopic}, using the same unsupervised training and evaluation splits to ensure fair comparison. Unlike our method, SciTopic uses pseudo-labels during training, which introduces weak supervision; therefore, we gray out its entry in the results for clarity. As shown in Tab.~\ref{tab:model_comparison}, our method RDR achieves the best performance across both datasets, with an accuracy of 84.86 on AG News and 52.91 on 20 News Groups. RDR also leads in NMI (61.66 and 56.57) and ARI (65.24 and 39.96), outperforming all fully unsupervised baselines, and even surpassing the pseudo-supervised SciTopic model.
\appendix
\vspace{-5pt}
\section{Appendix}
This appendix presents the complete results of our Real Deep Research (RDR) analysis across a wide range of domains. We include detailed domain-level surveys (e.g., AI, robotics, computer vision, natural language processing), perspective-based breakdowns (e.g., input/output modeling in foundation models, sensor/action perspectives in robotics), and trend analyses to track the evolution of research focus over time. These results collectively offer a structured and insightful view of the research landscape, serving as a valuable reference for both new and experienced researchers.

\begin{tabularx}{0.95\textwidth}{@{} X @{\dotfill} r @{}}
  \texttt{\textbf{Domain Survey}} & \pageref{tab:fm} \\
  \hspace*{2em} \texttt{Foundation Model} & \pageref{tab:fm} \\
  \hspace*{2em} \texttt{Robotics} & \pageref{tab:ro} \\
  \hspace*{2em} \texttt{Computer Vision} & \pageref{tab:cv} \\
  \hspace*{2em} \texttt{Natural Language Processing} & \pageref{tab:nlp} \\
  \hspace*{2em} \texttt{Machine Learning} & \pageref{tab:ml} \\
  \hspace*{2em} \texttt{Nature} & \pageref{tab:nature} \\
  \hspace*{2em} \texttt{Science} & \pageref{tab:science} \\
  \texttt{\textbf{Perspective Survey}} & \pageref{tab:fm-input} \\
  \hspace*{2em} \texttt{Foundation Model Perspectives} & \pageref{tab:fm-input} \\
  \hspace*{2em} \texttt{Robotic Perspectives} & \pageref{tab:ro-sensors} \\
  \texttt{\textbf{Trend Analysis}} & \pageref{tab:tr-cv} \\
  \hspace*{2em} \texttt{Computer Vision} & \pageref{tab:tr-cv} \\
  \hspace*{2em} \texttt{Robotics} & \pageref{tab:tr-robotic} \\
  \hspace*{2em} \texttt{Natural Language Processing} & \pageref{tab:tr-nlp} \\
  \hspace*{2em} \texttt{Machine Learning} & \pageref{tab:tr-ml-topics} \\
\end{tabularx}

\clearpage
% \section{Domain Survey}
% \label{sec:domain-survey}
\begin{table}[ht]
\centering\scriptsize      % one step smaller than \small
\begin{tabularx}{\textwidth}{
  @{} >{\raggedright\arraybackslash}p{0.02\linewidth}  % Category
      >{\raggedright\arraybackslash}p{0.23\linewidth}  % Sub-category
      >{\raggedright\arraybackslash}X                  % What is covered
      >{\raggedright\arraybackslash}p{0.27\linewidth}  % Examples
      >{\centering\arraybackslash}p{0.06\linewidth}@{}}% Cluster
% ----- COLOURED HEADER (safe from row-banding) -----
\toprule
\textbf{Cat.} & \textbf{Sub-category} & \textbf{What is covered} &
\textbf{Typical examples} & \textbf{\scriptsize Cluster} \\
\midrule

%=========== 1. Modalities & representations ===========
\rowcolor{CategoryBG}
\multicolumn{5}{@{}l}{\bfseries 1.\;Model modalities \& representations \cite{Chen2024Internvl,FengLi2025Llava,YiWang*2024Internvideo,ShengbangTong2024Cambrian,YechengWu2025Vila,MinShi2025Eagle,ShengqiongWu2025Towards,XueyanZou2024Interfacing,JiaboYe2025Mplug,HaoZhang*2024Llava}}\\
& 1.1 Vision–Language \cite{Chen2024Internvl,YechengWu2025Vila,WeihanWang2024Cogvlm,ChenyuZhou2025Learning,Chen2024Scaling}
& Foundation models that jointly process images/video and natural language.
& Vision-Language Models; Vision-Language Robotic FMs
& \scriptsize 0, 3 \\

\rowcolor{AltGray}
& 1.2 Multimodal (\(\ge 3\)) \cite{Srivastava2024Omnivec,XuZheng*2024Learning,ZhiyangXu2025Modality,XinyanChen2025X,XuZheng*2024Centering}
& Architectures/objectives agnostic to the exact mix of modalities.
& Multimodal Foundation Models; Multimodal LLMs; Multimodal Large Language Models
& \scriptsize 1, 4, 5, 6 \\

& 1.3 Open-vocabulary grounding \cite{DaveZhenyuChen2024Scanrefer,Margffoy-Tuay2024Dynamic,ZhenxiangLin2024Wildrefer,YufeiZhan2024Griffon,WeichengKuo2024Findit}
& Linking free-form text to modality-specific regions or anchors.
& Open-Vocabulary Grounding
& \scriptsize 1 \\
\bottomrule
\end{tabularx}
\caption{Domain Survey for Foundation Model.}
\label{tab:fm}
\end{table}
\begin{table}[ht]
\centering\scriptsize      % one step smaller than \small
\begin{tabularx}{\textwidth}{
  @{} >{\raggedright\arraybackslash}p{0.02\linewidth}  % Category
      >{\raggedright\arraybackslash}p{0.23\linewidth}  % Sub-category
      >{\raggedright\arraybackslash}X                  % What is covered
      >{\raggedright\arraybackslash}p{0.34\linewidth}  % Examples
      >{\centering\arraybackslash}p{0.06\linewidth}@{}}% Cluster
\toprule
\textbf{Cat.} & \textbf{Sub-category} & \textbf{What is covered} &
\textbf{Typical examples} & \textbf{\scriptsize Cluster} \\
\midrule

%=========== 1. Perception & Mapping ===========
\rowcolor{CategoryBG}
\multicolumn{5}{@{}l}{\bfseries 1.\;Perception \& Mapping \cite{AbbasSadat2024Perceive,JuexiaoZhang2024Multiview,WenzhaoZheng2024Occworld,MedhiniNarasimhan2024Seeing,Harwath2024Jointly}}\\
& 1.1 Multimodal sensor fusion \cite{Liang2024Deep,Yin2024Fusion,TimBroedermann2024Muses,Zimmer2024Tumtraf,TsunHsuanWang2024V}
& Fuse heterogeneous sensors for richer scene understanding
& LiDAR–Camera Fusion; Radar–Camera Fusion; V2X Cooperative Perception ...
& \scriptsize 0,6,7,8,14,16 \\

\rowcolor{AltGray}
& 1.2 3D reconstruct/occupancy \cite{WenzhaoZheng2024Occworld,Huang2024Selfocc,HaisongLiu2024Fully,WeiyangLiu2024Structural,Peng2024Learning}
& Build dense or sparse geometric maps for localisation
& 3-D SLAM \& Reconstruction; 3-D Occupancy; Efficient 3-D Representation
& \scriptsize 0,8,16 \\

& 1.3 BEV / top-view mapping \cite{Zhao2024Improving,Yin2024Fusion,Chambon2024Pointbev,KaiJiang2024Da,ZhiqiLi2024Bevformer}
& Bird’s-eye or top-down representations for planning
& BEV Perception; V2X Collaborative Perception
& \scriptsize 0,14,16 \\

%=========== 2. Manipulation & Grasping ===========
\rowcolor{CategoryBG}
\multicolumn{5}{@{}l}{\bfseries 2.\;Manipulation \& Grasping \cite{OmidTaheri2024Grab,SamarthBrahmbhatt2024Contactpose,YanWu2024Saga,DylanTurpin2024Grasp,YuanchenJu2024Robo}}\\
\bottomrule
\end{tabularx}
\caption{Domain Survey for Robotics.}
\label{tab:ro}
\end{table}
\begin{table}[ht]
\vspace{-40pt}
\centering\scriptsize
\begin{tabularx}{\textwidth}{
@{} >{\raggedright\arraybackslash}p{0.07\linewidth}
>{\raggedright\arraybackslash}p{0.18\linewidth}
>{\raggedright\arraybackslash}X
>{\raggedright\arraybackslash}p{0.28\linewidth}
>{\centering\arraybackslash}p{0.08\linewidth}@{}}
\toprule
\textbf{Category} & \textbf{Sub-category} & \textbf{What is covered} &
\textbf{Typical examples} & \textbf{\scriptsize Cluster} \\
\midrule

\rowcolor{CategoryBG}
\multicolumn{5}{@{}l}{\bfseries 1. Robust \& Generalizable Learning} \\
& 1.1 Adversarial / OOD Robustness
& Defending against or detecting malicious, anomalous or out-of-distribution inputs.
& adversarial robustness; deepfake detection; anomaly detection; out-of-distribution detection
& \scriptsize 0,2,5 \\
\rowcolor{AltGray}
& 1.2 Domain Adaptation \& Generalization
& Transferring models across different domains, devices or persons without performance drop.
& domain adaptation; domain generalization; test-time adaptation; person re-identification
& \scriptsize 3,6 \\
& 1.3 Low-Data Learning
& Learning reliably from scarce, imbalanced or continually arriving data.
& few-shot; continual learning; long-tailed recognition; federated learning
& \scriptsize 0,1,2 \\

\rowcolor{CategoryBG}
\multicolumn{5}{@{}l}{\bfseries 2. Representation \& Model Efficiency} \\
& 2.1 Representation Learning \& Distillation
& Un/semisupervised learning and distillation techniques that build informative, explainable features.
& self-supervised learning; semi-supervised segmentation; pseudo-label consistency; representation learning; knowledge distillation; explainability
& \scriptsize 4,5,7,10 \\
\rowcolor{AltGray}
& 2.2 Efficient Architectures
& Designing compact, hardware-friendly or automatically searched neural networks.
& efficient ViT; neural architecture compression; sparse/quantized NAS
& \scriptsize 9 \\
\bottomrule
\end{tabularx}
\caption{Domain Survey for Computer Vision.}
\label{tab:cv}
\end{table}
\begin{table}[ht]
\vspace{-60pt}
\centering\scriptsize % one step smaller than \small
\begin{tabularx}{\textwidth}{
@{} >{\raggedright\arraybackslash}p{0.06\linewidth} % Category
>{\raggedright\arraybackslash}p{0.18\linewidth}   % Sub-category
>{\raggedright\arraybackslash}X                   % What is covered
>{\raggedright\arraybackslash}p{0.26\linewidth}   % Examples
>{\centering\arraybackslash}p{0.05\linewidth}@{}} % Cluster
\toprule
\textbf{Category} & \textbf{Sub-category} & \textbf{What is covered} &
\textbf{Typical examples} & \textbf{\scriptsize Cluster} \\
\midrule

\rowcolor{CategoryBG}
\multicolumn{5}{@{}l}{\bfseries 1.\;Information Extraction}\\
& 1.1 Entity \& Relation Extraction
& Automatic detection of named entities plus the semantic relations or events connecting them.
& Named Entity Recognition, Relation Extraction, Event Extraction
& \scriptsize 0 \\

\rowcolor{CategoryBG}
\multicolumn{5}{@{}l}{\bfseries 2.\;Text Generation \& Summarization}\\
& 2.1 Summarization \& Keyphrase Generation
& Producing concise summaries or keyphrases from longer documents.
& Summarization, Keyphrase, Evaluation
& \scriptsize 12 \\
\rowcolor{AltGray}
& 2.2 Controllable \& Stylistic Generation
& Generating text under user-specified style or attribute constraints.
& Style transfer, controllable text generation, representations
& \scriptsize 27 \\

\rowcolor{CategoryBG}
\multicolumn{5}{@{}l}{\bfseries 3.\;Dialogue \& Conversational Systems}\\
& 3.1 Task-oriented Dialogue
& Dialogue systems that track state and generate responses to accomplish user goals.
& Dialogue systems, Response generation, Dialogue state tracking
& \scriptsize 14 \\
\bottomrule
\end{tabularx}
\caption{Domain Survey for Natural Language Processing.}
\label{tab:nlp}
\end{table}
\begin{table}[ht]
\vspace{-60pt}
\centering\scriptsize      % one step smaller than \small
\begin{tabularx}{\textwidth}{
@{} >{\raggedright\arraybackslash}p{0.06\linewidth}   % Category
>{\raggedright\arraybackslash}p{0.18\linewidth}       % Sub-category
>{\raggedright\arraybackslash}X                       % What is covered
>{\raggedright\arraybackslash}p{0.3\linewidth}       % Examples
>{\centering\arraybackslash}p{0.05\linewidth}@{}}     % Cluster
\toprule
\textbf{Category} & \textbf{Sub-category} & \textbf{What is covered} &
\textbf{Typical examples} & \textbf{\scriptsize Cluster} \\
\midrule

% ---------- Category 1 ----------
\rowcolor{CategoryBG}
\multicolumn{5}{@{}l}{\bfseries 1.\;Generative Modelling \& Media Synthesis}\\
& 1.1 Image / Video Generation \& Editing
& Models that create or edit 2-D or temporal visual content via generative techniques.
& Generative modeling; Image synthesis/editing; Diffusion-based methods; Optimal Transport; Diffusion Models
& \scriptsize 1, 3 \\
\rowcolor{AltGray}
& 1.2 3-D Object \& Molecule Generation
& Generating 3-D shapes or molecular structures using geometry-aware or equivariant models.
& 3D shape generation; neural implicit representations; point cloud reconstruction; equivariant GNNs; 3D molecular generation; drug discovery
& \scriptsize 9, 16 \\
& 1.3 Audio \& Speech Synthesis
& Producing speech or audio from text or multimodal cues, via diffusion models.
& text-to-speech; audio-visual; diffusion
& \scriptsize 2 \\

% ---------- Category 2 ----------
\rowcolor{CategoryBG}
\multicolumn{5}{@{}l}{\bfseries 2.\;Representation \& Transfer Learning}\\
& 2.1 Continual / Few-Shot / Domain Adaptation
& Adapting models continually, with few examples, or across shifting domains.
& Continual Learning; Few-Shot Learning; Domain Adaptation
& \scriptsize 0 \\
\bottomrule
\end{tabularx}
\caption{Domain Survey for Machine Learning.}
\label{tab:ml}
\end{table}
\begin{table}[ht]
\centering\scriptsize
\begin{tabularx}{\textwidth}{
  @{} >{\raggedright\arraybackslash}p{0.06\linewidth}  % Category
      >{\raggedright\arraybackslash}p{0.2\linewidth}  % Sub-category
      >{\raggedright\arraybackslash}X                  % What is covered
      >{\raggedright\arraybackslash}p{0.28\linewidth}  % Examples
      >{\centering\arraybackslash}p{0.06\linewidth}@{}}% Cluster
\toprule
\textbf{Category} & \textbf{Sub-category} & \textbf{What is covered} &
\textbf{Typical examples} & \textbf{\scriptsize Cluster} \\
\midrule

%=========== 1. Life Sciences & Biomedicine ===========
\rowcolor{CategoryBG}
\multicolumn{5}{@{}l}{\bfseries 1.\;Life Sciences \& Biomedicine}\\
& 1.1 Immuno-oncology \& Metabolic Signalling
& Immune mechanisms in cancer and metabolic cues that modulate them
& T-cell immunity; tumour microenvironment; metabolic signalling
& \scriptsize 0 \\

\rowcolor{AltGray}
& 1.2 Cancer Genomics \& Epigenetics
& Genetic and epigenetic alterations driving oncogenesis and therapy response
& Cancer; DNA repair; epigenetics
& \scriptsize 1 \\

& 1.3 Infectious Disease \& Microbiome Interactions
& Host–pathogen dynamics and microbiome ecology shaping antimicrobial strategies
& Host–pathogen interactions; antimicrobial therapeutics; microbiome dynamics
& \scriptsize 2 \\

\rowcolor{AltGray}
& 1.4 Neuro-immune Metabolism \& Aging
& Crosstalk between immune system, metabolism and brain across aging
& Neuroimmunology; metabolism; aging
& \scriptsize 3 \\

& 1.5 Genome Editing \& Microbial/Plant Immunity
& Engineering genomes and decoding microbial & plant defence mechanisms
\& CRISPR-based genome editing; bacterial anti-phage defence; plant immune signalling
& \scriptsize 4 \\

\rowcolor{AltGray}
& 1.6 Protein \& RNA Engineering
& Designing proteins and regulating chromatin/RNA to control cell function
& Protein design; chromatin regulation; RNA processing
& \scriptsize 5 \\
\bottomrule
\end{tabularx}
\caption{Domain Survey for Natural related Topics.}
\label{tab:nature}
\end{table}

\begin{table}[ht]
\centering\scriptsize      % one step smaller than \small
\begin{tabularx}{\textwidth}{
  @{} >{\raggedright\arraybackslash}p{0.10\linewidth}  % Category
      >{\raggedright\arraybackslash}p{0.18\linewidth}  % Sub-category
      >{\raggedright\arraybackslash}X                  % What is covered
      >{\raggedright\arraybackslash}p{0.29\linewidth}  % Examples
      >{\centering\arraybackslash}p{0.10\linewidth}@{}}% Cluster
\toprule
\textbf{Category} & \textbf{Sub-category} & \textbf{What is covered} &
\textbf{Typical examples} & \textbf{\scriptsize Cluster} \\
\midrule

\rowcolor{CategoryBG}
\multicolumn{5}{@{}l}{\bfseries 1.\;Earth \& Environmental Sciences}\\
& 1.1 Climate \& Ecosystem Dynamics
& Interactions among climate change, carbon cycling and biodiversity, including conservation responses
& Biodiversity loss; Conservation strategies; Climate change impacts; Climate change; Carbon cycle; Environmental impacts
& \scriptsize 0,1 \\
\rowcolor{AltGray}
& 1.2 Geophysical Processes
& Physical processes shaping Earth’s solid and cryospheric systems
& Earthquakes; Volcanism; Ice dynamics
& \scriptsize 2 \\

\rowcolor{CategoryBG}
\multicolumn{5}{@{}l}{\bfseries 2.\;Space Science}\\
& 2.1 Stellar \& Space-Plasma Physics
& Physics of stars, solar activity and the interstellar medium
& Compact objects; Interstellar medium; Solar activity
& \scriptsize 9 \\

\rowcolor{CategoryBG}
\multicolumn{5}{@{}l}{\bfseries 3.\;Biological Sciences}\\
& 3.1 Evolutionary Genomics
& Genetic mechanisms driving adaptation and speciation
& Evolutionary genomics; adaptation; speciation
& \scriptsize 3 \\
\bottomrule
\end{tabularx}
\caption{Structured overview of clustered science research areas.}
\label{tab:science_cluster_overview}
\end{table}
\begin{table}[ht]
\centering\scriptsize
\begin{tabularx}{\textwidth}{
@{} >{\raggedright\arraybackslash}p{0.10\linewidth}  % Category
>{\raggedright\arraybackslash}p{0.18\linewidth}      % Sub-category
>{\raggedright\arraybackslash}X                      % What is covered
>{\raggedright\arraybackslash}p{0.26\linewidth}      % Examples
>{\centering\arraybackslash}p{0.10\linewidth}@{}}    % Cluster
\toprule
\textbf{Category} & \textbf{Sub-category} & \textbf{What is covered} & \textbf{Typical examples} & \textbf{\scriptsize Cluster} \\
\midrule

\rowcolor{CategoryBG}
\multicolumn{5}{@{}l}{\bfseries 1. Pharmacogenomics \& Genetics-Guided Therapy}\\
& 1.1 Cytochrome P450 Genotype-Driven Therapy
& Links CYP450 genetic variants to drug exposure and response for individualised dosing.
& CYP2C19 pharmacogenetics, star-allele variability, precision antithrombotic therapy, antidepressant pharmacogenetics
& \scriptsize 1,3,4 \\
\rowcolor{AltGray}
& 1.2 Transporter Pharmacogenetics
& Examines genetic variation in drug transporters and its impact on safety and efficacy.
& SLCO1B1 variants, statin-associated muscle symptoms
& \scriptsize 6 \\
& 1.3 PGx Implementation \& Economic Evaluation
& Assesses clinical decision support, workflow integration, and the cost-effectiveness of pharmacogenomics.
& pharmacogenomics implementation, clinical decision support, cost-effectiveness
& \scriptsize 11 \\
\rowcolor{AltGray}
& 1.4 Oncology / High-Risk Therapy PGx
& Uses germline variants to predict toxicity and guide dosing of high-risk or anticancer drugs.
& DPYD variants, TPMT variants, NUDT15 variants, chemotherapy toxicity prediction, pharmacogenetic-guided dosing
& \scriptsize 21 \\

\rowcolor{CategoryBG}
\multicolumn{5}{@{}l}{\bfseries 2. Quantitative Pharmacology \& Model-Informed Drug Development}\\
\bottomrule
\end{tabularx}
\caption{Domain Survey for Science related Survey.}
\label{tab:science}
\end{table}

\clearpage
% \section{Perspective Survey}
% \label{sec:perspective-survey}
\begin{table}[ht]
\centering\scriptsize      % one step smaller than \small
\begin{tabularx}{\textwidth}{
  @{} >{\raggedright\arraybackslash}p{0.06\linewidth}  % Category
      >{\raggedright\arraybackslash}p{0.15\linewidth}  % Sub-category
      >{\raggedright\arraybackslash}X                  % What is covered
      >{\raggedright\arraybackslash}p{0.32\linewidth}  % Examples
      >{\centering\arraybackslash}p{0.06\linewidth}@{}}% Cluster
% ----- COLOURED HEADER (safe from row-banding) -----
\toprule
\textbf{Category} & \textbf{Sub-category} & \textbf{What is covered} &
\textbf{Typical examples} & \textbf{\scriptsize Cluster} \\
\midrule

%=========== 1. Textual inputs ===========
\rowcolor{CategoryBG}
\multicolumn{5}{@{}l}{\bfseries 1.\;Textual inputs}\\
& 1.1 Large-scale tokenized corpora
& Massive general-domain text for LM pre-training
& Web pages; Wikipedia; books; C4; Pile; WikiText; OpenWebText; SlimPajama
& \scriptsize 11 \\

\rowcolor{AltGray}
& 1.2 Prompt \& interaction data
& User/system prompts and model replies gathered for alignment, RLHF or robustness
& Prompts/questions; model responses; preference/reward labels; adversarial triggers; long-context demonstrations
& \scriptsize 0, 2 \\

& 1.3 Problem statements with context
& Natural-language tasks paired with explicit structured knowledge or code/data schemas
& NL problem + knowledge graph/database schema/code stub; reasoning traces or step-by-step solutions
& \scriptsize 14 \\

%=========== 2. Visual inputs (images) ===========
\rowcolor{CategoryBG}
\multicolumn{5}{@{}l}{\bfseries 2.\;Visual inputs (images)}\\
& 2.1 Raw images
& Canonical labelled/unlabelled images after basic augmentation
& ImageNet, CIFAR, COCO photos; medical scans
& \scriptsize 19 \\

\rowcolor{AltGray}
& 2.2 Cued images
& Images supplied with auxiliary spatial/sensor cues
& Low-light or blurry photos + masks; camera poses; depth/event data; points/boxes
& \scriptsize 17 \\
\bottomrule
\end{tabularx}
\caption{Structured summary of input types used in foundation-model papers.}
\label{tab:fm-input}
\end{table}

\begin{table}[ht]
\centering\scriptsize      % one step smaller than \small
\begin{tabularx}{\textwidth}{
  @{} >{\raggedright\arraybackslash}p{0.10\linewidth}  % Category
      >{\raggedright\arraybackslash}p{0.18\linewidth}  % Sub-category
      >{\raggedright\arraybackslash}X                  % What is covered
      >{\raggedright\arraybackslash}p{0.29\linewidth}  % Examples
      >{\centering\arraybackslash}p{0.10\linewidth}@{}}% Cluster
\toprule
\textbf{Category} & \textbf{Sub-category} & \textbf{What is covered} &
\textbf{Typical examples} & \textbf{\scriptsize Cluster} \\
\midrule

%=========== 1. Representation & Architecture ===========
\rowcolor{CategoryBG}
\multicolumn{5}{@{}l}{\bfseries 1.\;Representation \& Architecture}\\
& 1.1 Token \& latent representation
& Mapping raw data to discrete/continuous tokens or latents
& Latent representation learning; token/patch embedding; visual-token projection … 
& \scriptsize 0, 14, 17, 18, 19 \\

\rowcolor{AltGray}
& 1.2 Attention \& Transformer variants
& Architectural changes that make attention cheaper or deeper
& Sparse/low-rank attention; spatiotemporal attention; positional scaling …
& \scriptsize 2, 10, 11, 14, 17, 18, 19 \\

& 1.3 Mixture-of-Experts \& routing
& Dynamic selection of expert blocks or routes
& Modular MoE; dynamic routing; MoE token routing …
& \scriptsize 0, 11, 15, 17, 19 \\

%=========== 2. Generative Paradigms ===========
\rowcolor{CategoryBG}
\multicolumn{5}{@{}l}{\bfseries 2.\;Generative Paradigms}\\
& 2.1 Diffusion \& score-based generation
& Noise-to-data generative flows
& UNet diffusion; latent diffusion; guided conditional sampling …
& \scriptsize 2, 3, 4, 5, 10, 13, 18 \\

\rowcolor{AltGray}
& 2.2 Energy-based \& control formulations
& Sampling by minimising energy or solving control processes
& Energy-based score learning; optimal-control SDE; solver-accelerated inversion …
& \scriptsize 13 \\
\bottomrule
\end{tabularx}
\caption{Structured summary of modeling techniques used in foundation-model papers.}
\label{tab:fm-modeling}
\end{table}

\begin{table}[ht]
\centering\scriptsize
\begin{tabularx}{\textwidth}{
  @{} >{\raggedright\arraybackslash}p{0.07\linewidth}
      >{\raggedright\arraybackslash}p{0.2\linewidth}
      >{\raggedright\arraybackslash}X
      >{\raggedright\arraybackslash}p{0.34\linewidth}
      >{\centering\arraybackslash}p{0.06\linewidth}@{}}
\toprule
\textbf{Category} & \textbf{Sub-category} & \textbf{What is covered} &
\textbf{Typical examples} & \textbf{\scriptsize Cluster} \\
\midrule

\rowcolor{CategoryBG}
\multicolumn{5}{@{}l}{\bfseries 1.\;Language-centric outputs}\\
& 1.1 Token probabilities \& sequences
& Autoregressive LMs: token logits or generated text
& next-token probability distributions; generated token sequences
& \scriptsize 2 \\

\rowcolor{AltGray}
& 1.2 Aligned LLM responses
& Instruction-tuned completions for reasoning, safety, long-context
& helpful-harmless-truthful responses; safe refusals; reasoning-enhanced; hallucination-reduced
& \scriptsize 3, 15 \\

& 1.3 Reasoning traces \& answers
& Chain-of-thought steps plus final decoded result
& chain-of-thought reasoning; intermediate outputs; final answers
& \scriptsize 0 \\

\rowcolor{AltGray}
& 1.4 Visually-grounded NL outputs
& Language grounded in image/video content
& captions; visual-QA; reasoning with bounding boxes, masks
& \scriptsize 9 \\

\rowcolor{CategoryBG}
\multicolumn{5}{@{}l}{\bfseries 2.\;Generative visual \& multimodal outputs}\\
& 2.1 Photorealistic images
& High-fidelity images from text or prompts
& photorealistic; identity-preserving; context-coherent images; text-conditioned images
& \scriptsize 1, 18 \\
\bottomrule
\end{tabularx}
\caption{Structured summary of output types used in foundation-model papers.}
\label{tab:fm-output}
\end{table}
\begin{table}[ht]
\centering\scriptsize
\begin{tabularx}{\textwidth}{
  @{} >{\raggedright\arraybackslash}p{0.08\linewidth}
      >{\raggedright\arraybackslash}p{0.18\linewidth}
      >{\raggedright\arraybackslash}X
      >{\raggedright\arraybackslash}p{0.3\linewidth}
      >{\centering\arraybackslash}p{0.08\linewidth}@{}}
\toprule
\textbf{Category} & \textbf{Sub-category} & \textbf{What is covered} &
\textbf{Typical examples} & \textbf{\scriptsize Cluster} \\
\midrule

%=========== 1. Language-modeling ===========
\rowcolor{CategoryBG}
\multicolumn{5}{@{}l}{\bfseries 1.\;Language-modeling objectives}\\
& Next / Masked-token Prediction
& Minimize CE on next/masked token.
& Next-/masked-token pred.; LM; CE/NLL min.; aux reg.
& \scriptsize 4 \\

\rowcolor{AltGray}
& General LLM Advancement
& Improve reasoning, alignment, efficiency, robustness.
& Reasoning; alignment; eval.; efficiency; robustness; multi-domain
& \scriptsize 12 \\

%=========== 2. Alignment & safety ===========
\rowcolor{CategoryBG}
\multicolumn{5}{@{}l}{\bfseries 2.\;Alignment \& safety objectives}\\
& Human-Preference Alignment
& Maximize learned reward; limit divergence.
& Pref. align; reward max.; safety-divergence reg.
& \scriptsize 1 \\

\rowcolor{AltGray}
& Hallucination \& Bias Mitigation
& Cut hallucinations/bias via grounding alignment.
& Hallucination det./mit.; x-modal align/ground; bias red.
& \scriptsize 0 \\

& General Safety \& Robustness
& Losses for safety, explainability, robust autonomy.
& Alignment; safety; efficiency; general.; explain.; autonomy
& \scriptsize 6 \\
\bottomrule
\end{tabularx}
\caption{Structured summary of learning objectives used in foundation-model papers.}
\label{tab:fm-objective}

\end{table}

\begin{table}[ht]
\centering\scriptsize
\begin{tabularx}{\textwidth}{
  @{} >{\raggedright\arraybackslash}p{0.07\linewidth}
      >{\raggedright\arraybackslash}p{0.24\linewidth}
      >{\raggedright\arraybackslash}X
      >{\raggedright\arraybackslash}p{0.27\linewidth}
      >{\centering\arraybackslash}p{0.07\linewidth}@{}}
\toprule
\textbf{Category} & \textbf{Sub-category} & \textbf{What is covered} &
\textbf{Typical examples} & \textbf{\scriptsize Cluster} \\
\midrule

%=========== 1. Pre-Training \& Representation Learning ===========
\rowcolor{CategoryBG}
\multicolumn{5}{@{}l}{\bfseries 1.\;Pre-Training \& Representation Learning}\\
& 1.1 Contrastive/masked vision--language pre-training
& Learn aligned image--text embeddings before any task-specific tuning.
& ViT/CLIP contrastive/masked pretrain; strong aug.; $T{=}0.07$; 40--600\,ep finetune.
& \scriptsize 5 \\
\rowcolor{AltGray}
& 1.2 Adapter-aided diffusion image/video pre-training
& Freeze released checkpoints; add lightweight adapters to scale.
& Frozen ckpt + LoRA/prompt; AdamW + cos LR; prog-res; CF guidance.
& \scriptsize 13 \\

%=========== 2. Fine-Tuning \& Adaptation ===========
\rowcolor{CategoryBG}
\multicolumn{5}{@{}l}{\bfseries 2.\;Fine-Tuning \& Adaptation}\\
& 2.1 Vision--language instruction tuning
& Turn a frozen VLM into an instruction follower.
& Image--text pretrain$\to$inst. tune; PEFT; opt. RLHF.
& \scriptsize 1 \\
\rowcolor{AltGray}
& 2.2 Parameter-efficient domain adaptation
& Keep backbone frozen; adapt via prompts/adapters only.
& Prompt/adapter/LoRA; distill or contrastive shift.
& \scriptsize 10 \\
& 2.3 Instruction SFT + retrieval alignment
& Align an LLM with retrieval and preferences.
& Multi-stage SFT; retrieval ctx; DPO/RLHF; rerank$\to$generate.
& \scriptsize 6 \\
\bottomrule
\end{tabularx}
\caption{Structured summary of training recipes used in foundation-model papers.}
\label{tab:fm-recipes}
\end{table}

\begin{table}[ht]
\centering\scriptsize      % one step smaller than \small
\begin{tabularx}{\textwidth}{
  @{} >{\raggedright\arraybackslash}p{0.1\linewidth}  % Category (unchanged)
      >{\raggedright\arraybackslash}p{0.18\linewidth}  % Sub-category – moved slightly left and narrower
      >{\raggedright\arraybackslash}X                  % What is covered – gets the largest share
      >{\raggedright\arraybackslash}p{0.22\linewidth}  % Examples – now smaller
      >{\centering\arraybackslash}p{0.10\linewidth}@{}}% Cluster
\toprule
\textbf{Category} & \textbf{Sub-category} & \textbf{What is covered} &
\textbf{Typical examples} & \textbf{\scriptsize Cluster} \\
\midrule

%=========== 1. Vision & Imaging Sensors ===========
\rowcolor{CategoryBG}
% \cite{XuanlinLi2024Evaluating,TsungWeiKe2024D,XiangyuWang2025Towards,ShuaihangYuan2024Gamap,SuvirMirchandani2024So}
\multicolumn{5}{@{}l}{\bfseries 1.\;Vision \& Imaging Sensors }
\\
& 1.1 RGB cameras
& Monocular, stereo, multi-view, surround-view or panoramic cameras producing color frames; used for appearance-based perception.
& front/side/rear vehicle cameras, egocentric/wrist/-head cameras, aerial/on-board cameras
& \scriptsize 0, 1, 2, 3, 4, 6, 7, 9, 10, 11, 13, 14, 15, 16, 17, 18, 19 \\

\rowcolor{AltGray}
& 1.2 RGB-D cameras
& Active or structured-light cameras that output synchronized color + depth images.
& Intel RealSense, Azure Kinect, panoramic RGB-D rigs
& \scriptsize 3, 7, 10, 11, 13, 14, 17, 18, 19 \\

& 1.3 Event (neuromorphic) cameras
& Asynchronous sensors emitting per-pixel brightness changes with micro-second latency.
& DVS, DAVIS
& \scriptsize 9 \\

\rowcolor{AltGray}
& 1.4 Thermal / LWIR cameras
& Passive long-wave IR imagers for temperature or night-vision cues.
& Thermal cameras, LWIR DoFP polarization cameras
& \scriptsize 3, 14 \\

%=========== 2. Depth & Range Sensors ===========
\rowcolor{CategoryBG}
% \cite{AnushriDixit2024Perceive,TsungWeiKe2024D,ShuaihangYuan2024Gamap,XiangyuWang2025Towards,ZiwenZhuang2024Humanoid}
\multicolumn{5}{@{}l}{\bfseries 2.\;Depth \& Range Sensors }\\
\bottomrule
\end{tabularx}
\caption{Structured summary of input sensors used in robotic papers.}
\label{tab:ro-sensors}
\end{table}

\begin{table}[ht]
\centering\scriptsize      % one step smaller than \small
\begin{tabularx}{\textwidth}{
  @{} >{\raggedright\arraybackslash}p{0.10\linewidth}  % Category
      >{\raggedright\arraybackslash}p{0.18\linewidth}  % Sub-category
      >{\raggedright\arraybackslash}X                  % What is covered
      >{\raggedright\arraybackslash}p{0.29\linewidth}  % Examples
      >{\centering\arraybackslash}p{0.10\linewidth}@{}}% Cluster
% ----- COLOURED HEADER (safe from row-banding) -----
\toprule
\textbf{Category} & \textbf{Sub-category} & \textbf{What is covered} &
\textbf{Typical examples} & \textbf{\scriptsize Cluster} \\
\midrule

%=========== 1. Ground-based mobile robots ===========
\rowcolor{CategoryBG}
% \cite{RiaDoshi2024Scaling,YuxingLong2024Instructnav,JingyunYang2024Equibot,NoriakiHirose2024Lelan,KuoHaoZeng2024Poliformer}
\multicolumn{5}{@{}l}{\bfseries 1.\;Ground-based mobile robots 
}\\
& 1.1 Small RC / off-road vehicles
& 1/10-scale cars, ATVs, skid-steer rovers for field tests
& RC cars/ATVs; small off-road vehicles
& \scriptsize 19 \\

\rowcolor{AltGray}
& 1.2 Kinematic vehicle models
& Bicycle/unicycle point-mass models (simulation-only)
& Simulated vehicle agents (kinematic/dynamic)
& \scriptsize 0 \\

%=========== 2. Aerial robots ===========
\rowcolor{CategoryBG}
\multicolumn{5}{@{}l}{\bfseries 2.\;Aerial robots }\\
& 2.1 Quadrotors / drones
& Four-rotor UAVs with cameras, LiDAR, IMU
& Quadrotor UAVs; drones
& \scriptsize 11, 19 \\

%=========== 3. Legged & humanoid robots ===========
\rowcolor{CategoryBG}
\multicolumn{5}{@{}l}{\bfseries 3.\;Legged \& humanoid robots}\\
& 3.1 Simulated legged agents
& Classic MuJoCo bodies for RL locomotion
& Hopper; HalfCheetah; Walker2d; Ant; Quadruped
& \scriptsize 11 \\
\bottomrule
\end{tabularx}
\caption{Structured summary of physical bodies used in robotic papers.}
\label{tab:ro-body}
\end{table}
\begin{table}[ht]
\centering\scriptsize      % one step smaller than \small
\begin{tabularx}{\textwidth}{
  @{} >{\raggedright\arraybackslash}p{0.10\linewidth}  % Category
      >{\raggedright\arraybackslash}p{0.18\linewidth}  % Sub-category
      >{\raggedright\arraybackslash}X                  % What is covered
      >{\raggedright\arraybackslash}p{0.29\linewidth}  % Examples
      >{\centering\arraybackslash}p{0.10\linewidth}@{}}% Cluster
\toprule
\textbf{Category} & \textbf{Sub-category} & \textbf{What is covered} &
\textbf{Typical examples} & \textbf{\scriptsize Cluster} \\
\midrule

%=========== 1. Direct joint-level outputs ===========
\rowcolor{CategoryBG}
\multicolumn{5}{@{}l}{\bfseries 1.\;Direct joint-level outputs}\\
& 1.1 Joint state read-outs
& Instantaneous articulated joint positions, orientations, angles, velocities
& “joint positions \& orientations”; “joint angles”; “joint velocities”; “mesh deformations”
& \scriptsize 0, 1, 9 \\

\rowcolor{AltGray}
& 1.2 Joint command signals
& Low-level motor targets (torque / position / velocity) that drive joint motion
& “joint torque/position commands”; “continuous motor control signals”; “PD control torques”
& \scriptsize 7, 11, 12, 16, 17 \\

& 1.3 Joint motion trajectories
& Time-indexed sequences of joint states the robot follows
& “motion sequences over time”; “planned joint trajectories”; “optimised 6-DoF trajectories”
& \scriptsize 0, 1, 17 \\

%=========== 2. Rigid-body / end-effector pose outputs ===========
\rowcolor{CategoryBG}
\multicolumn{5}{@{}l}{\bfseries 2.\;Rigid-body / end-effector pose outputs}\\
& 2.1 6-DoF body poses
& Position + orientation of whole robots, cameras or objects
& “6-DoF poses”; “rigid transformations”; “UAV 3-D position \& orientation”
& \scriptsize 6, 10, 16 \\

\rowcolor{AltGray}
& 2.2 End-effector pose + gripper
& Cartesian pose of manipulator tip plus gripper open/close state
& “6-DoF end-effector pose $(x,y,z,r,p,y)$”; “\texttt{gripper\_state} (open/close)”
& \scriptsize 4 \\
\bottomrule
\end{tabularx}
\caption{Structured summary of joint outputs used in robotic papers.}
\label{tab:ro-joint}
\end{table}

\begin{table}[ht]
\centering\scriptsize      % one step smaller than \small
\begin{tabularx}{\textwidth}{
  @{} >{\raggedright\arraybackslash}p{0.1\linewidth}  % Category
      >{\raggedright\arraybackslash}p{0.18\linewidth}  % Sub-category
      >{\raggedright\arraybackslash}X                  % What is covered
      >{\raggedright\arraybackslash}p{0.29\linewidth}  % Examples
      >{\centering\arraybackslash}p{0.10\linewidth}@{}}% Cluster
\toprule
\textbf{Category} & \textbf{Sub-category} & \textbf{What is covered} &
\textbf{Typical examples} & \textbf{\scriptsize Cluster} \\
\midrule

%=========== 1. Continuous Low-Level Actuation ===========
\rowcolor{CategoryBG}
\multicolumn{5}{@{}l}{\bfseries 1.\;Continuous Low-Level Actuation}\\
& 1.1 Joint-space commands
& Direct numerical inputs to individual joints or actuators, bounded by hardware limits.
& joint torques/positions/velocities; high-dimensional joint commands; bounded control inputs; finger-joint configs; parametrised joint trajectories
& \scriptsize 0, 4, 6, 10, 12, 14, 18 \\

\rowcolor{AltGray}
& 1.2 Vehicle / body dynamics commands
& Low-level controls that change a mobile base, ground-vehicle or aerial body state.
& steering angle; throttle / acceleration; braking; linear \& angular velocity; body-rate thrust; speed/direction for locomotion; lane-keeping
& \scriptsize 0, 1, 7, 10, 12, 13, 15 \\

%=========== 2. Mid-Level Pose \& Trajectory Control ===========
\rowcolor{CategoryBG}
\multicolumn{5}{@{}l}{\bfseries 2.\;Mid-Level Pose \& Trajectory Control}\\
& 2.1 End-effector \& gripper pose
& 6-DoF goals and time-parameterised trajectories for arms, grippers or aerial manipulators.
& continuous 6-DoF poses; pose deltas ($\Delta x$, $\Delta y$, $\Delta z$, $\Delta\text{roll}$, $\Delta\text{pitch}$, $\Delta\text{yaw}$); gripper open/close; gripper width/force; grasp trajectories
& \scriptsize 2, 6, 9, 10, 12, 14, 18 \\
\bottomrule
\end{tabularx}
\caption{Structured summary of action space used in robotic papers.}
\label{tab:ro-action}
\end{table}

\begin{table}[ht]
\centering\scriptsize      % one step smaller than \small
\begin{tabularx}{\textwidth}{
  @{} >{\raggedright\arraybackslash}p{0.1\linewidth}  % Category
      >{\raggedright\arraybackslash}p{0.18\linewidth}  % Sub-category
      >{\raggedright\arraybackslash}X                  % What is covered
      >{\raggedright\arraybackslash}p{0.22\linewidth}  % Examples
      >{\centering\arraybackslash}p{0.10\linewidth}@{}}% Cluster
\toprule
\textbf{Category} & \textbf{Sub-category} & \textbf{What is covered} &
\textbf{Typical examples} & \textbf{\scriptsize Cluster} \\
\midrule

%=========== 1. Autonomous-driving & Mobile-vehicle scenes ===========
\rowcolor{CategoryBG}
\multicolumn{5}{@{}l}{\bfseries 1.\;Autonomous-driving \& Mobile-vehicle scenes}\\
& 1.1 On-road urban / suburban / rural driving
& Real or simulated road networks with traffic, road rules, and weather variation.
& urban roads; highways; intersections; traffic lights/signs; \dots
& \scriptsize 1, 2, 6, 9, 12, 13, 19 \\

\rowcolor{AltGray}
& 1.2 Off-road, cross-country \& planetary terrain
& Structured or unstructured natural terrains requiring ground-robot locomotion.
& uneven ground; sand; gravel; snow; \dots
& \scriptsize 11 \\

%=========== 2. Manipulation workspaces ===========
\rowcolor{CategoryBG}
\multicolumn{5}{@{}l}{\bfseries 2.\;Manipulation workspaces}\\
& 2.1 Basic household tabletop
& Small cluttered indoor bench for reach-scale manipulation.
& cluttered tabletop; household objects; articulated fixtures; \dots
& \scriptsize 0 \\

\rowcolor{AltGray}
& 2.2 Kitchen \& household benchmark suites
& Standardised kitchen/tabletop scenes from RLBench, Meta-World, FrankaKitchen, Habitat, Ravens, etc.
& kitchen counters; RLBench station; FrankaKitchen; \dots
& \scriptsize 14, 17 \\

& 2.3 Assembly \& insertion tables
& Contact-rich assembly surfaces with precisely shaped parts.
& assembly workspace; peg-hole joints; plug-socket joints; \dots
& \scriptsize 18 \\
\bottomrule
\end{tabularx}
\caption{Structured summary of environment used in robotic papers.}
\label{tab:per-env}
\end{table}

\clearpage
% \section{Trend Analysis}
\label{sec:trend}
\begin{figure*}[htbp]      % * = full-width; remove it if not desired
    \centering
    \tiny
    \renewcommand{\arraystretch}{1.4}
    \setlength{\tabcolsep}{4pt}
    \vspace{-10pt}
    % ------------------------------------------------------------------
    \begin{tabular}{|p{0.30\textwidth}|p{0.30\textwidth}|p{0.30\textwidth}|}
    \hline
    % --------------------------- Row 1 -------------------------------
    \begin{minipage}[t]{\linewidth}\centering
        \includegraphics[width=0.9\linewidth]{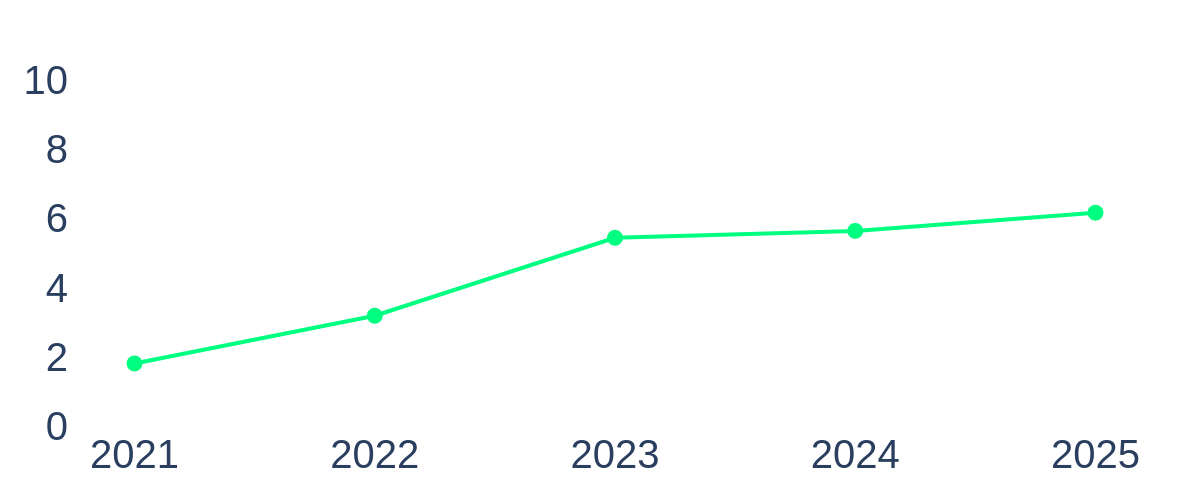}\par
        1.\;Vision-Language Pretraining, Zero-Shot Learning, Cross-Modal Retrieval
    \end{minipage} &
    \begin{minipage}[t]{\linewidth}\centering
        \includegraphics[width=0.9\linewidth]{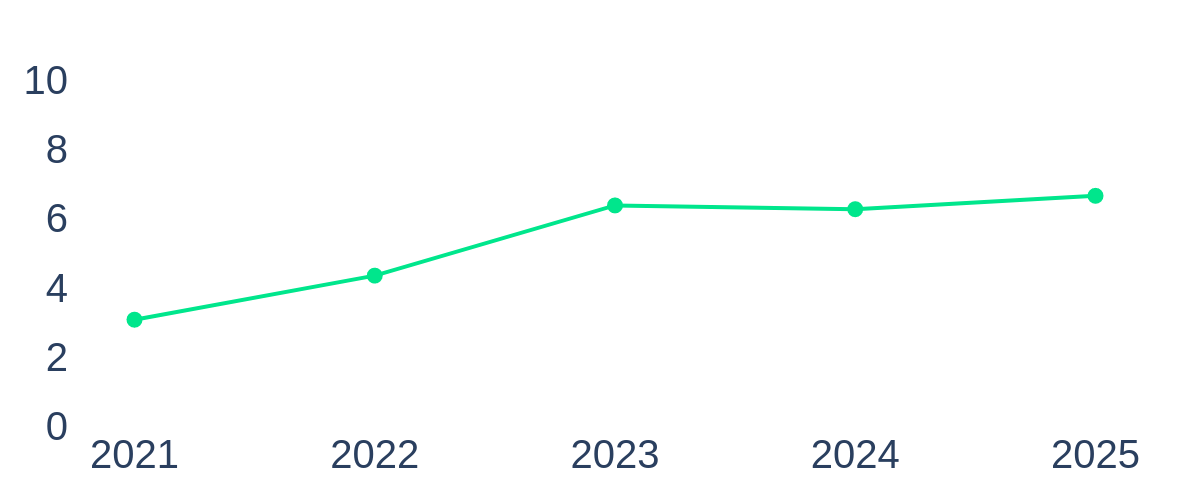}\par
        2.\;Neural Radiance Fields, 3D Gaussian Splatting, Dynamic Scene Reconstruction
    \end{minipage} &
    \begin{minipage}[t]{\linewidth}\centering
        \includegraphics[width=0.9\linewidth]{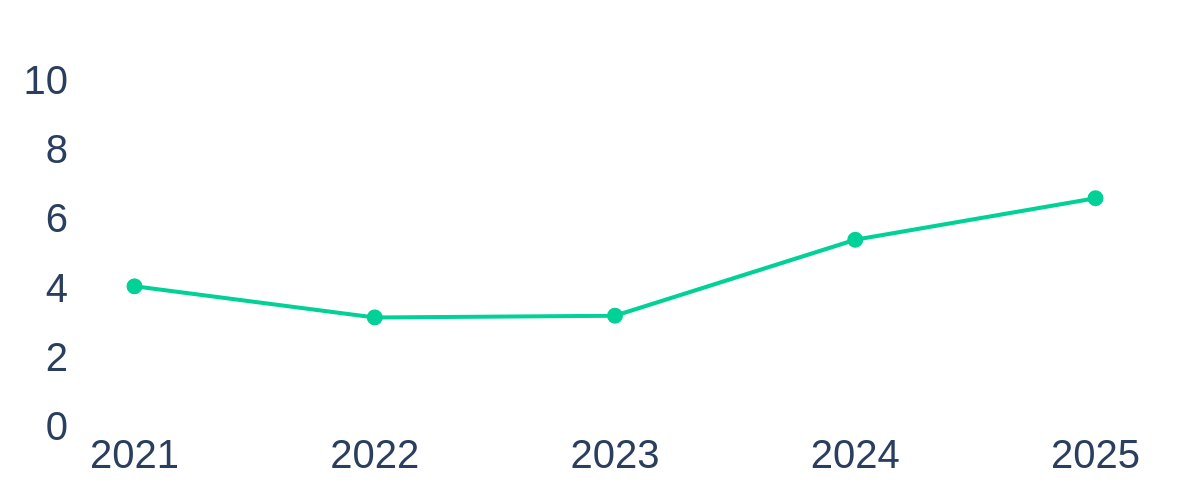}\par
        3.\;Neural Style Transfer, Diffusion-Based Generation, Controllable Editing
    \end{minipage}\\ \hline
    % --------------------------- Row 2 -------------------------------
    \begin{minipage}[t]{\linewidth}\centering
        \includegraphics[width=0.9\linewidth]{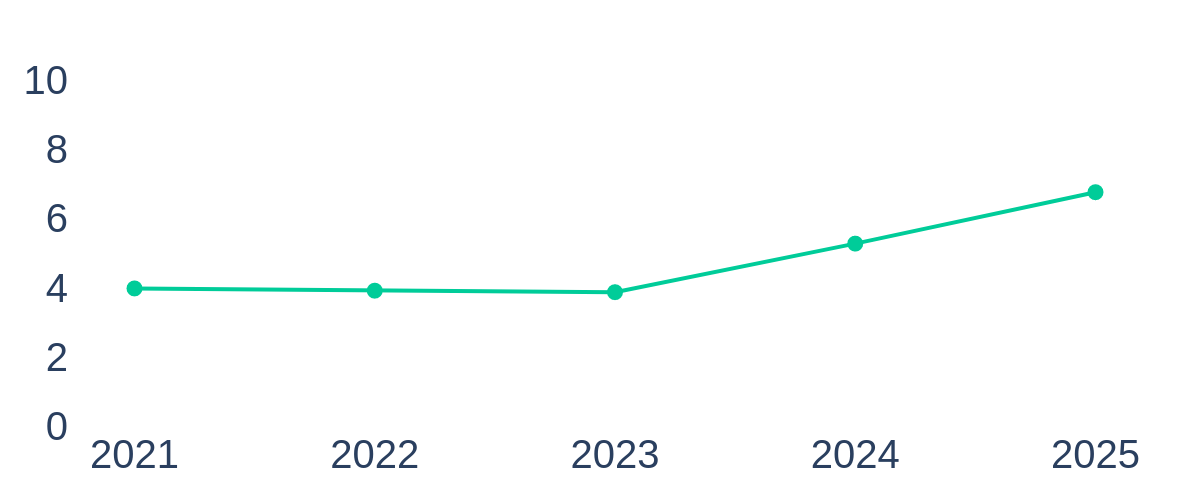}\par
        4.\;GAN, Image Inpainting/Translation, Disentangled Representation
    \end{minipage} &
    \begin{minipage}[t]{\linewidth}\centering
        \includegraphics[width=0.9\linewidth]{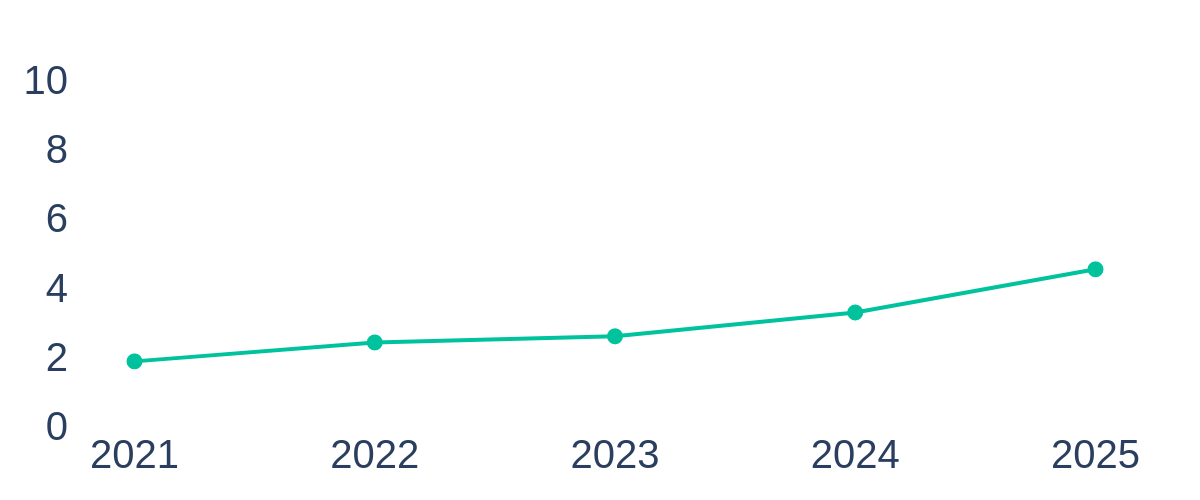}\par
        5.\;Embodied Navigation, 3D Vision-Language Grounding, Scene Graph Generation
    \end{minipage} &
    \begin{minipage}[t]{\linewidth}\centering
        \includegraphics[width=0.9\linewidth]{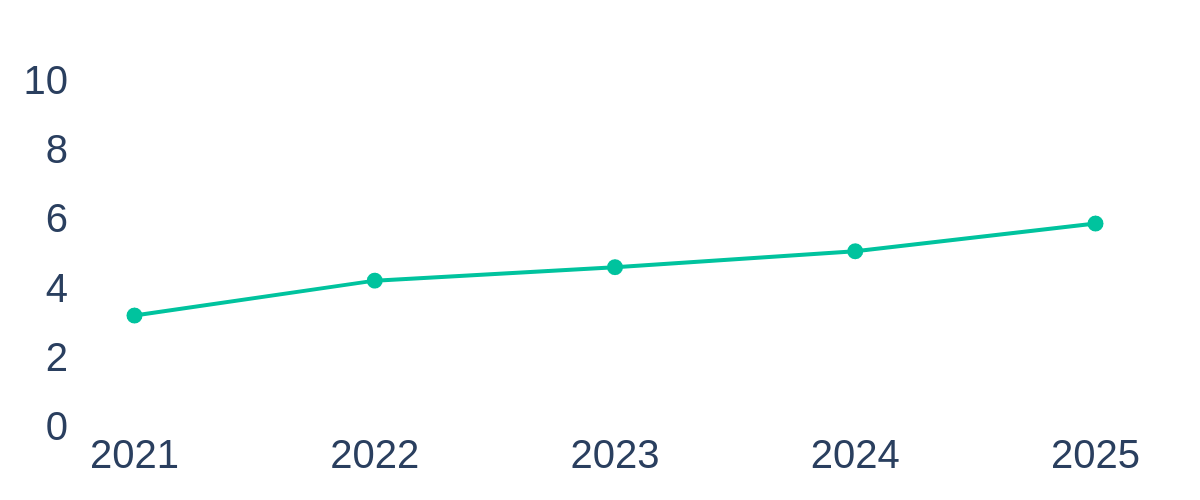}\par
        6.\;Adversarial Robustness, Federated Learning, Deepfake Detection
    \end{minipage}\\
\hline
    \end{tabular}
    \caption{Trend Visualization of Computer Vision Research}
    \label{tab:tr-cv}
\end{figure*}
\begin{figure*}[htbp]
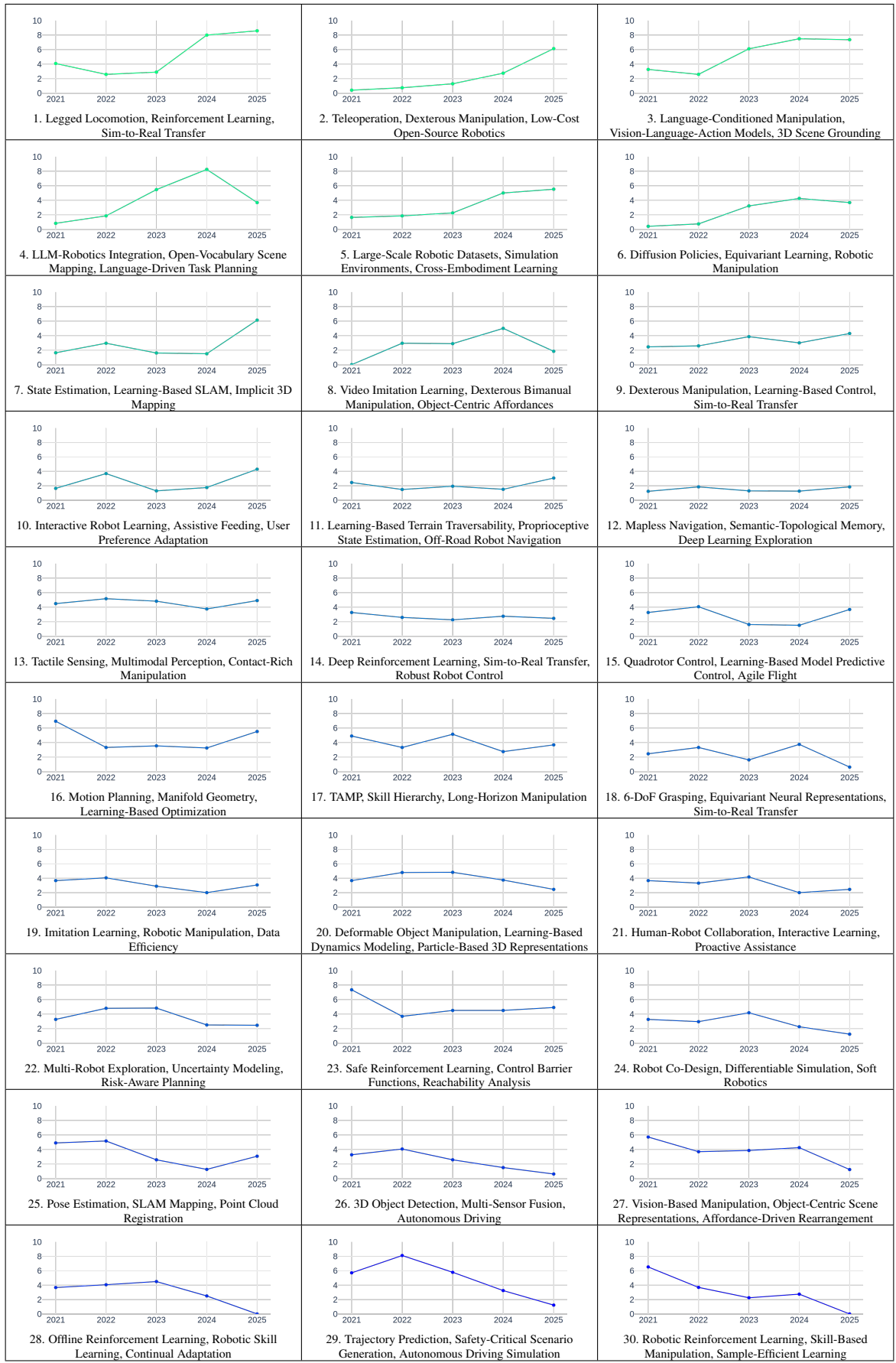
      % * = full-width; remove it if not desired
    \centering
    \tiny
    % Better spacing in the table
    \renewcommand{\arraystretch}{1.4}
    \setlength{\tabcolsep}{4pt}
    \vspace{-20pt}
    % ------------------------------------------------------------------
    \begin{tabular}{|p{0.30\textwidth}|p{0.30\textwidth}|p{0.30\textwidth}|}
    \hline
    % --------------------------- Row 1 -------------------------------
    \begin{minipage}[t]{\linewidth}\centering
        \includegraphics[width=0.9\linewidth]{Figures/rss_corl/0_legged_locomotion,_reinforcement_learning,_sim-to-real_transfer.png}\par
        1.\;Legged Locomotion, Reinforcement Learning, Sim-to-Real Transfer
    \end{minipage} &
    \begin{minipage}[t]{\linewidth}\centering
        \includegraphics[width=0.9\linewidth]{Figures/rss_corl/1_teleoperation,_dexterous_manipulation,_low-cost_open-source_robotics.png}\par
        2.\;Teleoperation, Dexterous Manipulation, Low-Cost Open-Source Robotics
    \end{minipage} &
    \begin{minipage}[t]{\linewidth}\centering
        \includegraphics[width=0.9\linewidth]{Figures/rss_corl/2_language-conditioned_manipulation,_vision-language-action_models,_3d_scene_grounding.png}\par
        3.\;Language-Conditioned Manipulation, Vision-Language-Action Models, 3D Scene Grounding
    \end{minipage}\\
\hline
    \end{tabular}
    \caption{Trend Visualization of Robotics Research}
    \label{tab:tr-robotic}
\end{figure*}
\begin{figure*}[htbp]      % * = full-width; remove it if not desired
    \centering
    \tiny
    % Better spacing in the table
    \renewcommand{\arraystretch}{1.4}
    \setlength{\tabcolsep}{4pt}
    \vspace{-20pt}
    % ------------------------------------------------------------------
    \begin{tabular}{|p{0.30\textwidth}|p{0.30\textwidth}|p{0.30\textwidth}|}
    \hline
    % --------------------------- Row 1 -------------------------------
    \begin{minipage}[t]{\linewidth}\centering
        \includegraphics[width=0.9\linewidth]{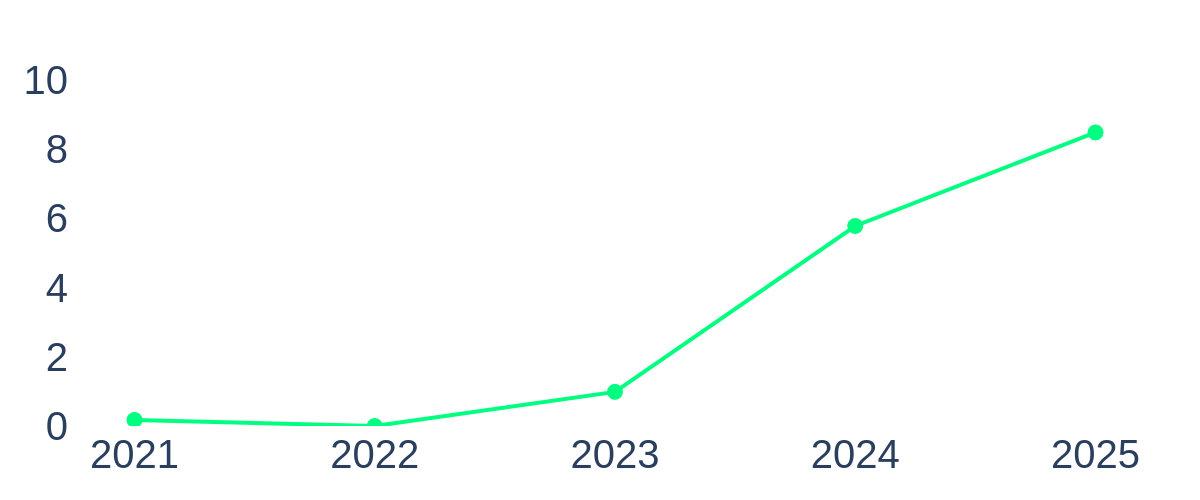}\par
        1.\;LLM Evaluation, Alignment Methods, Human Feedback
    \end{minipage} &
    \begin{minipage}[t]{\linewidth}\centering
        \includegraphics[width=0.9\linewidth]{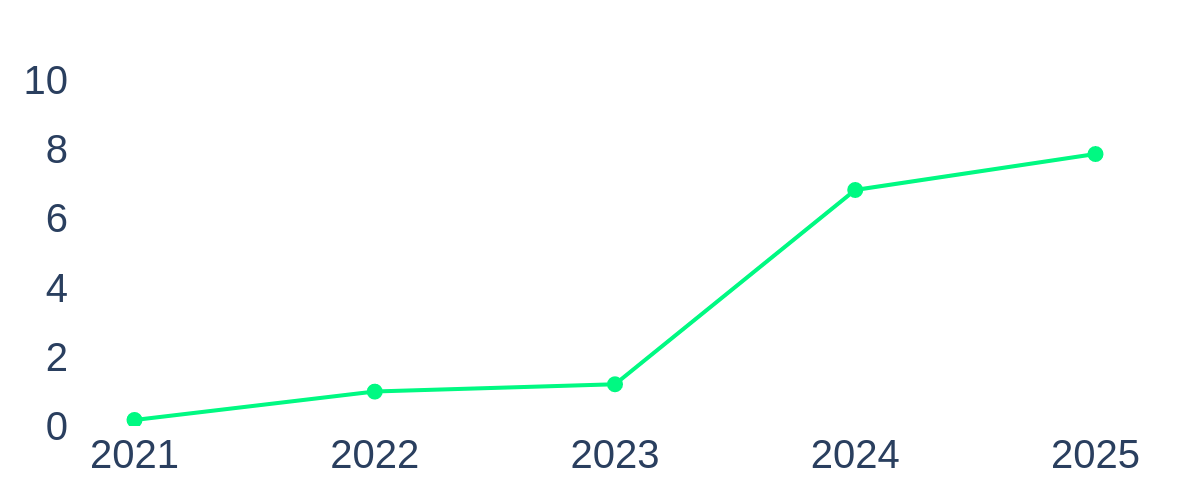}\par
        2.\;LLM Agents, Interactive Planning, Theory of Mind
    \end{minipage} &
    \begin{minipage}[t]{\linewidth}\centering
        \includegraphics[width=0.9\linewidth]{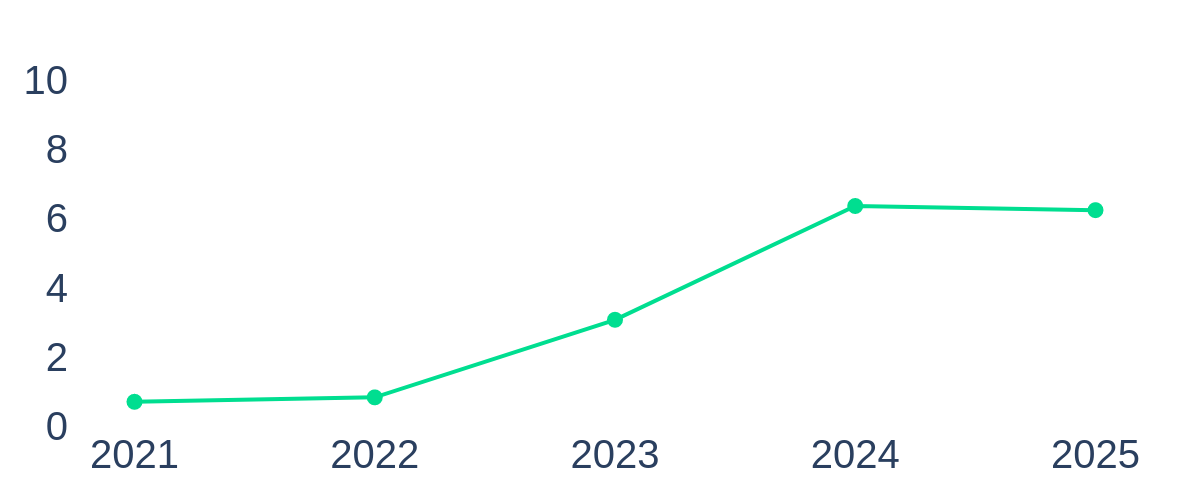}\par
        3.\;Large Language Models, Mathematical Reasoning, Chain-of-Thought Prompting
    \end{minipage}\\ \hline
    % --------------------------- Row 2 -------------------------------
    \begin{minipage}[t]{\linewidth}\centering
        \includegraphics[width=0.9\linewidth]{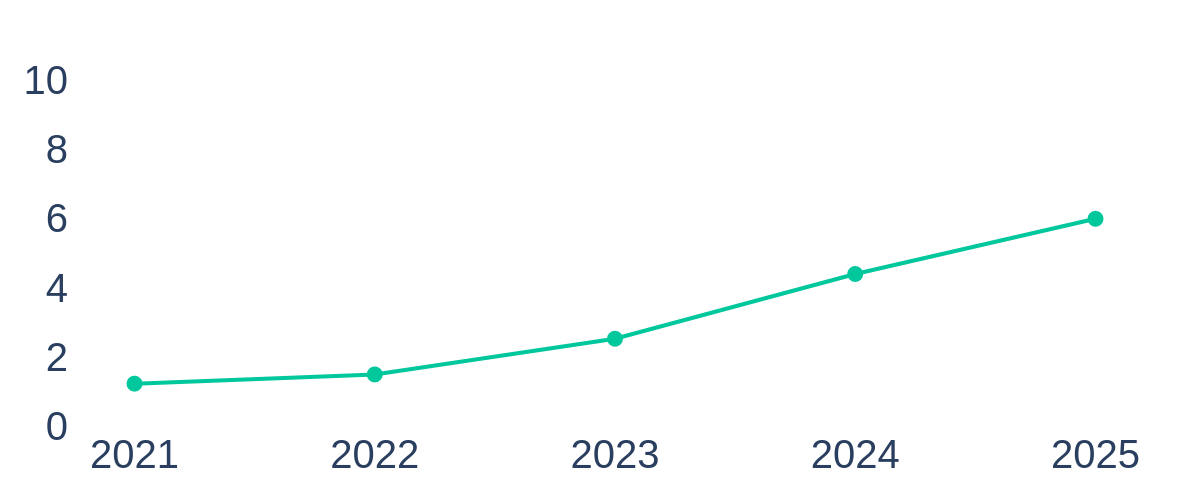}\par
        4.\;Adversarial Robustness, Backdoor Attacks, Privacy Preservation
    \end{minipage} &
    \begin{minipage}[t]{\linewidth}\centering
        \includegraphics[width=0.9\linewidth]{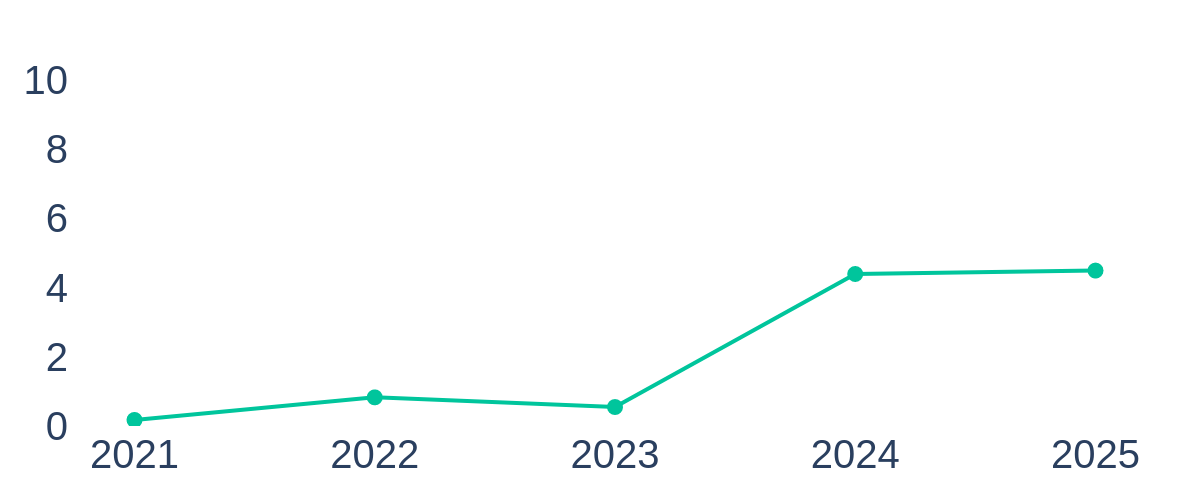}\par
        5.\;Hallucination, Knowledge Editing, Calibration
    \end{minipage} &
    \begin{minipage}[t]{\linewidth}\centering
        \includegraphics[width=0.9\linewidth]{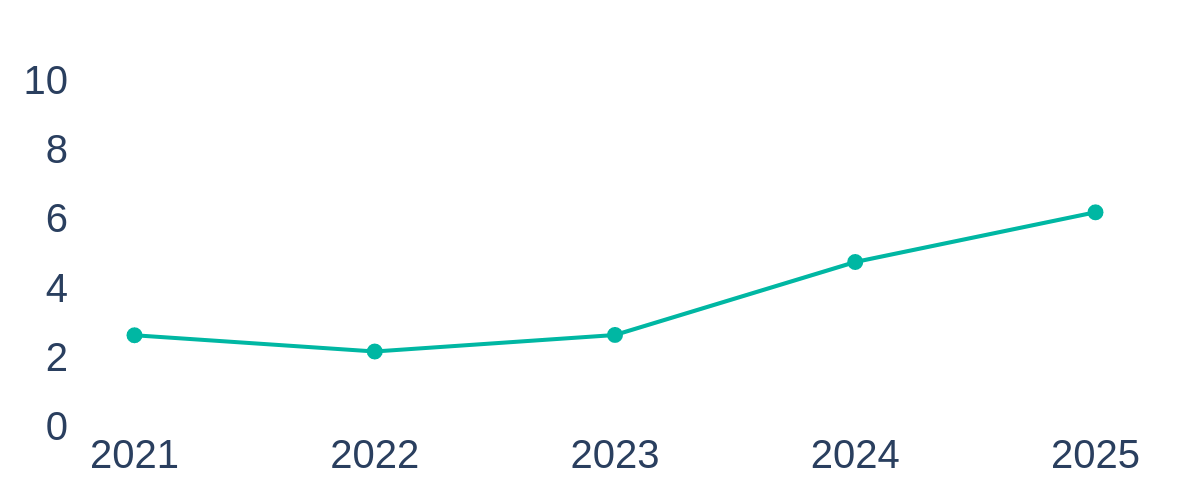}\par
        6.\;Dense Retrieval, Open-Domain Question Answering, Retrieval-Augmented Generation
    \end{minipage}\\
\hline
    \end{tabular}
    \caption{Trend Visualization of NLP Research}
    \label{tab:tr-nlp}
\end{figure*}
\begin{figure*}[htbp]      % * = full-width; remove it if not desired
    \centering
    \tiny
    % Better spacing in the table
    \renewcommand{\arraystretch}{1.4}
    \setlength{\tabcolsep}{4pt}
    \vspace{-20pt}
    % ------------------------------------------------------------------
    \begin{tabular}{|p{0.30\textwidth}|p{0.30\textwidth}|p{0.30\textwidth}|}
    \hline
    % --------------------------- Row 1 -------------------------------
    \begin{minipage}[t]{\linewidth}\centering
        \includegraphics[width=0.9\linewidth]{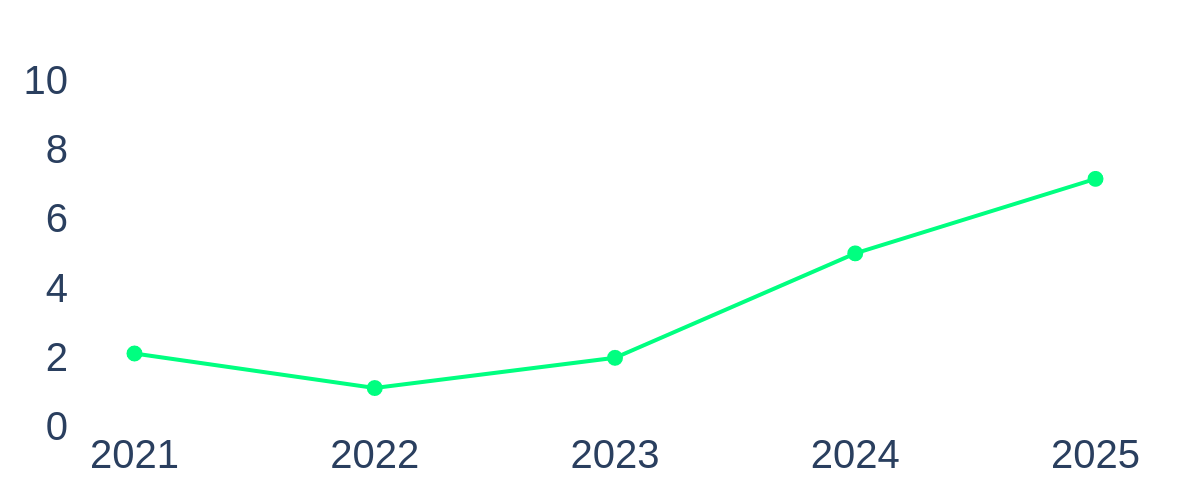}\par
        1.\;Program Synthesis, Code Generation, Theorem Proving
    \end{minipage} &
    \begin{minipage}[t]{\linewidth}\centering
        \includegraphics[width=0.9\linewidth]{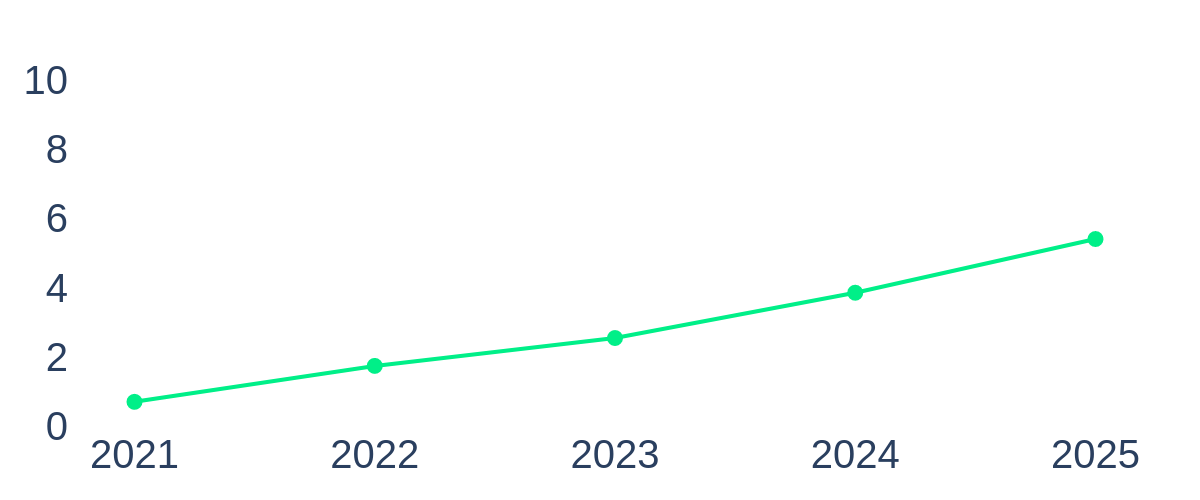}\par
        2.\;Vision-Language Reasoning, Knowledge Graph Learning, Compositional Generalization
    \end{minipage} &
    \begin{minipage}[t]{\linewidth}\centering
        \includegraphics[width=0.9\linewidth]{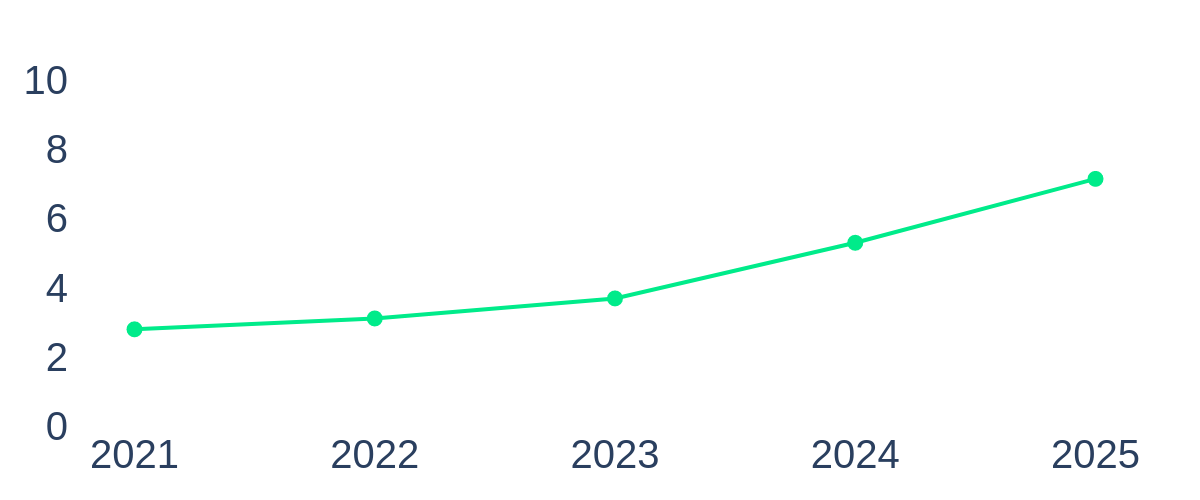}\par
        3.\;Language Modeling, Retrieval Augmentation, Representation Learning
    \end{minipage}\\ \hline
    % --------------------------- Row 2 -------------------------------
    \begin{minipage}[t]{\linewidth}\centering
        \includegraphics[width=0.9\linewidth]{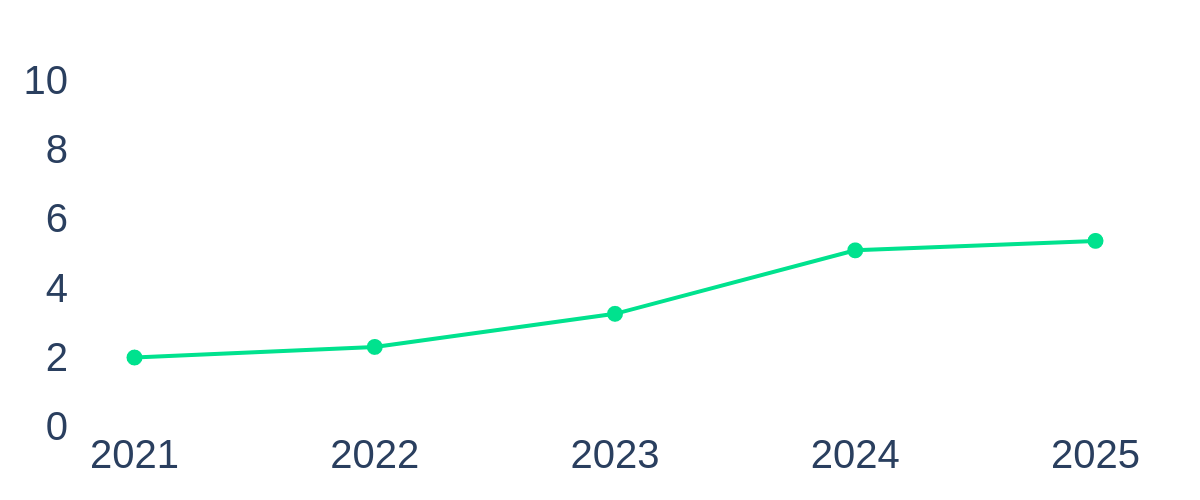}\par
        4.\;Generative Modeling, Image Synthesis \& Editing, Diffusion-Based Methods
    \end{minipage} &
    \begin{minipage}[t]{\linewidth}\centering
        \includegraphics[width=0.9\linewidth]{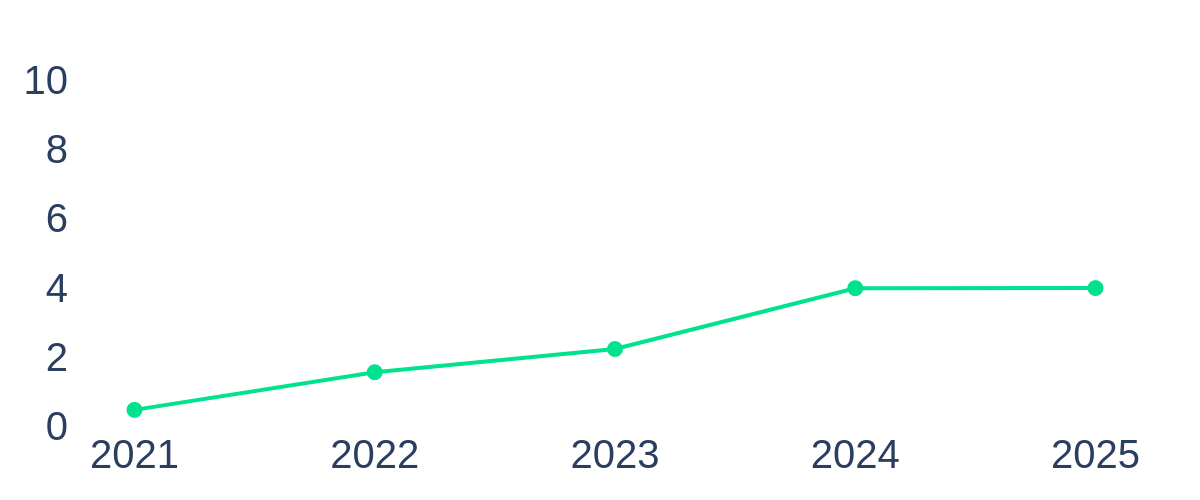}\par
        5.\;3D Shape Generation, Neural Implicit Representations, Point-Cloud Reconstruction
    \end{minipage} &
    \begin{minipage}[t]{\linewidth}\centering
        \includegraphics[width=0.9\linewidth]{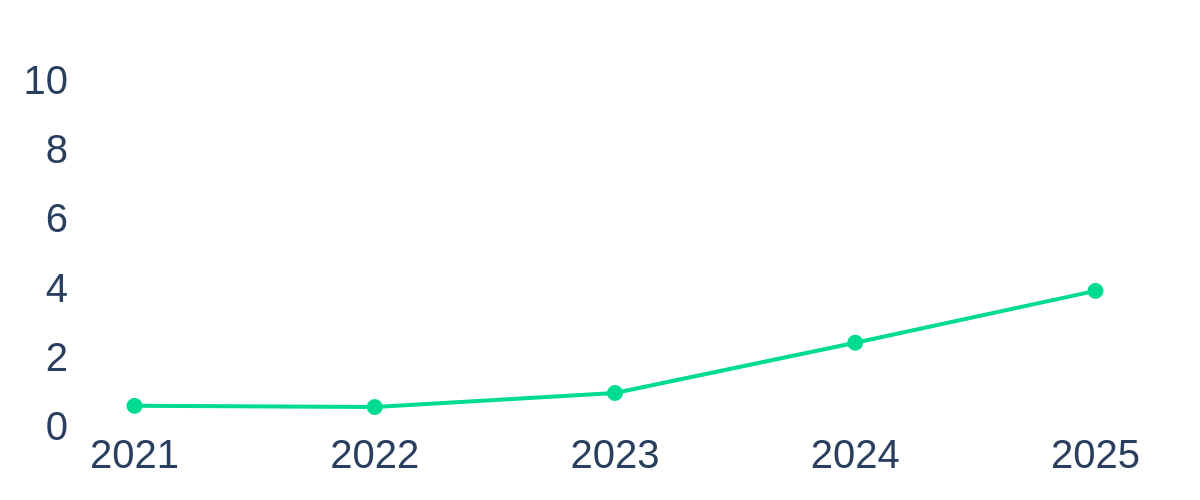}\par
        6.\;Dialogue Systems, Multi-Agent Collaboration, Reinforcement Learning
    \end{minipage}\\
\hline
    \end{tabular}
    \caption{Trend Visualization of Machine Learning Research}
    \label{tab:tr-ml-topics}
\end{figure*}

\clearpage
\bibliography{main}
\end{document}